\documentclass[10pt,journal]{IEEEtran}
\usepackage[table]{xcolor}

\ifCLASSOPTIONcompsoc
  \usepackage[nocompress]{cite}
\else
  \usepackage{cite}
\fi
\ifCLASSINFOpdf
\else
\fi
\usepackage{xr}
\makeatletter
\newcommand*{\addFileDependency}[1]{
	\typeout{(#1)}
	\@addtofilelist{#1}
	\IfFileExists{#1}{}{\typeout{No file #1.}}
}
\makeatother

\usepackage{float}
\usepackage{calc}

\usepackage{hyperref} 
\usepackage{graphics,color,wrapfig}
\usepackage{url}

\usepackage{breakurl}
\usepackage{color}
\usepackage{amsthm}
\usepackage{amsmath,amssymb}
\usepackage{amsfonts}
\usepackage{array}
\newcolumntype{P}[1]{>{\centering\arraybackslash}m{#1}}
\usepackage[tmargin=0.5in,bmargin=0.5in,lmargin=0.63in,rmargin=0.63in]{geometry}
\usepackage{multirow}
\usepackage{graphicx}
\usepackage{standalone}
\usepackage{booktabs}
\usepackage{algpseudocode} 
\usepackage{algorithm2e} 
\let\oldnl\nl
\newcommand{\nonl}{\renewcommand{\nl}{\let\nl\oldnl}}
\usepackage{makecell}
\usepackage[bottom]{footmisc}
\usepackage{enumitem}
\theoremstyle{definition}
\usepackage{tikz}

\usepackage[justification=centering]{caption} 
\usepackage{subcaption}

\usepackage{mathtools}

\usepackage{bbm}
\usepackage{physics}

\makeatletter
\renewcommand{\@algocf@capt@plain}{above}
\makeatother
\RestyleAlgo{ruled}
\LinesNumbered

\newcommand{\calB}{\mathcal{B}}

\newcommand{\calW}{\mathcal{W}}
\newcommand{\calN}{\mathcal{N}}

\newcommand{\calG}{\mathcal{G}}

\newcommand{\calU}{\mathcal{U}}

\newcommand{\bfG}{\mathbf{G}}

\newcommand{\bfw}{\mathbf{w}}
\newcommand{\bfz}{\mathbf{z}}

\newtheorem{theorem}{Theorem}

\newtheorem{example}{Example}

\newtheorem{remark}{Remark}

{\begin{list}{}%
		{\setlength{\leftmargin}{#1}}%
		\item[]%
	}
	{\end{list}}

\newcounter{relctr} 
\everydisplay\expandafter{\the\everydisplay\setcounter{relctr}{0}} 

\AtBeginDocument{} 

\DeclareCaptionSubType * [alph]{table}
\def\*#1{\boldsymbol{\mathbf{#1}}}

\usepackage{trimclip}

\usepackage{pgf}
\usepackage{collcell}

\newcommand*{\MinNumber}{0}%
\newcommand*{\MaxNumber}{1}%

\newcommand{\ApplyGradient}[1]{%
	\pgfmathsetmacro{\PercentColor}{100.0*(#1-\MinNumber)/(\MaxNumber-\MinNumber)}%
	\edef\x{\noexpand\cellcolor{red!\PercentColor}}\x\textcolor{black}{#1}%
}
\newcolumntype{R}{>{\collectcell\ApplyGradient}{r}<{\endcollectcell}}

\begin{document}
%

\title{Detection and Mitigation of \\Byzantine Attacks in Distributed Training}
%
%
%
%

\author{Konstantinos~Konstantinidis, Namrata~Vaswani,~\IEEEmembership{Fellow,~IEEE}
        and~Aditya~Ramamoorthy,~\IEEEmembership{Senior~Member,~IEEE}
		\IEEEcompsocitemizethanks{
			\IEEEcompsocthanksitem This work was supported in part by the National Science Foundation (NSF) under grants CCF-1910840 and CCF-2115200. The material in this work has appeared in part at the 2022 IEEE International Symposium on Information Theory (ISIT). (Corresponding author: Aditya Ramamoorthy.)\protect
			\IEEEcompsocthanksitem The authors are with the Department of Electrical and Computer Engineering, Iowa State University, Ames, IA 50011, USA (e-mail: \{kostas, namrata, adityar\}@iastate.edu).
		}
}

%
%

\ifCLASSOPTIONpeerreview
	\markboth{Journal of \LaTeX\ Class Files,~Vol.~14, No.~8, August~2015}%
	{Shell \MakeLowercase{\textit{et al.}}: Bare Demo of IEEEtran.cls for Computer Society Journals}
\fi
%



\IEEEtitleabstractindextext{%

\begin{abstract}
A plethora of modern machine learning tasks require the utilization of large-scale distributed clusters as a critical component of the training pipeline. However, abnormal Byzantine behavior of the worker nodes can derail the training and compromise the quality of the inference. Such behavior can be attributed to unintentional system malfunctions or orchestrated attacks; as a result, some nodes may return arbitrary results to the parameter server (PS) that coordinates the training. Recent work considers a wide range of attack models and has explored robust aggregation and/or computational redundancy to correct the distorted gradients. 

In this work, we consider attack models ranging from strong ones: $q$ omniscient adversaries with full knowledge of the defense protocol that can change from iteration to iteration to weak ones: $q$ randomly chosen adversaries with limited collusion abilities which only change every few iterations at a time. Our algorithms rely on redundant task assignments coupled with detection of adversarial behavior. We also show the convergence of our method to the optimal point under common assumptions and settings considered in literature. For strong attacks, we demonstrate a reduction in the fraction of distorted gradients ranging from 16\%-99\% as compared to the prior state-of-the-art. Our top-1 classification accuracy results on the CIFAR-10 data set demonstrate 25\% advantage in accuracy (averaged over strong and weak scenarios) under the most sophisticated attacks compared to state-of-the-art methods.
\end{abstract}

}

\maketitle

\IEEEdisplaynontitleabstractindextext

%
\IEEEpeerreviewmaketitle

\section{Introduction and Background}
\label{sec:introduction}

%
%
%
%
\IEEEPARstart{I}ncreasingly complex machine learning models with large data set sizes are nowadays routinely trained on distributed clusters. A typical setup consists of a single central machine (\emph{parameter server} or PS) and multiple worker machines. The PS owns the data set, assigns gradient tasks to workers, and coordinates the protocol. The workers then compute gradients of the loss function with respect to the model parameters. These computations are returned to the PS, which \emph{aggregates} them, updates the model, and maintains the global copy of it. The new copy is communicated back to the workers. Multiple iterations of this process are performed until convergence has been achieved. PyTorch \cite{pytorch}, TensorFlow \cite{tensorflow}, MXNet \cite{mxnet}, CNTK \cite{CNTK} and other frameworks support this architecture.

These setups offer significant speedup benefits and enable training challenging, large-scale models. Nevertheless, they are vulnerable to misbehavior by the worker nodes, i.e., when a subset of them returns erroneous computations to the PS, either inadvertently or on purpose. This ``\emph{Byzantine}'' behavior can be attributed to a wide range of reasons. The principal causes of inadvertent errors are hardware and software malfunctions (e.g., \cite{flipping_bits_kim}). 
Reference \cite{bit_flip_attack_rakin} exposes the vulnerability of neural networks to such failures and identifies weight parameters that could maximize accuracy degradation. The gradients may also be distorted in an adversarial manner. As ML problems demand more resources, many jobs are often outsourced to external commodity servers (cloud) whose security cannot be guaranteed. Thus, an adversary may be able to gain control of some devices and fool the model. The distorted gradients can derail the optimization and lead to low test accuracy or slow convergence. 

Achieving robustness in the presence of Byzantine node behavior and devising training algorithms that can efficiently aggregate the gradients has inspired several works \cite{gupta_allerton_2019, aggregathor, ramchandran_saddle_point, ramchandran_optimal_rates, cong_generalized_sgd, blanchard_krum, ChenSX17, byzshield, lagrange_cdc, detox, draco, data_encoding}. The first idea is to filter the corrupted computations from the training without attempting to identify the Byzantine workers. Specifically, many existing papers use majority voting and median-based defenses \cite{gupta_allerton_2019, aggregathor, ramchandran_saddle_point, ramchandran_optimal_rates, cong_generalized_sgd, blanchard_krum, ChenSX17} for this purpose. In addition, several works also operate by replicating the gradient tasks \cite{byzshield, lagrange_cdc, detox, draco, data_encoding} allowing for consistency checks across the cluster. The second idea for mitigating Byzantine behavior involves detecting the corrupted devices and subsequently ignoring their calculations \cite{regatti2020bygars, zeno, alistarh_neurips_2018}, in some instances paired with redundancy \cite{draco}. In this work, we propose a technique that combines the usage of redundant tasks, filtering, and detection of Byzantine workers. Our work is applicable to a broad range of assumptions on the Byzantine behavior.

There is much variability in the adversarial assumptions that prior work considers. For instance, prior work differs in the maximum number of adversaries considered, their ability to collude, their possession of knowledge involving the data assignment and the protocol, and whether the adversarial machines are chosen at random or systematically. 
We will initially examine our methods under strong adversarial models similar to those in prior work \cite{aspis_isit, byzshield, alie, cong_generalized_sgd, ramchandran_optimal_rates, bulyan, auror}. We will then extend our algorithms to tackle weaker failures that are not necessarily adversarial but rather common in commodity machines \cite{flipping_bits_kim,bit_flip_attack_rakin,SIGNSGD}. We expand on related work in the upcoming Section \ref{sec:rel_work_contrib}.

\section{Related Work and Summary of Contributions}
\label{sec:rel_work_contrib}

\subsection{Related Work}
All work in this area (including ours) assumes a reliable parameter server that possesses the global data set and can assign specific subsets of it to workers. \emph{Robust aggregation} methods have also been proposed for federated learning \cite{robust_fedml_avestimehr, robust_fedml_jin}; however, as we make no assumption of privacy, our work, as well as the methods we compare with do not apply to federated learning.

One category of defenses splits the data set into $K$ \emph{batches} and assigns one to each worker with the ultimate goal of suitably aggregating the results from the workers. Early work in the area \cite{blanchard_krum} established that no \emph{linear aggregation} method (such as averaging) can be robust even to a single adversarial worker. This has inspired alternative methods collectively known as \emph{robust aggregation}. Majority voting, geometric median, and squared-distance-based techniques fall into this category \cite{aggregathor, ramchandran_saddle_point, ramchandran_optimal_rates, cong_generalized_sgd, blanchard_krum, ChenSX17}. 


One of the most popular robust aggregation techniques is known as \emph{mean-around-median} or \emph{trimmed mean} \cite{cong_generalized_sgd, ramchandran_optimal_rates}. It handles each dimension of the gradient separately and returns the average of a subset of the values that are closest to the median.
\emph{Auror} \cite{auror} is a variant of trimmed mean which partitions the values of each dimension into two clusters using \emph{k-means} and discards the smaller cluster if the distance between the two exceeds a threshold; the values of the larger cluster are then averaged.
\emph{signSGD} in \cite{SIGNSGD} transmits only the sign of the gradient vectors from the workers to the PS and exploits majority voting to decide the overall update; this practice reduces the communication time and denies any individual worker too much effect on the update.

\emph{Krum} in \cite{blanchard_krum} 
chooses a single honest worker for the next model update, discarding the data from the rest of them. 
The chosen gradient is the one closest to its $k\in\mathbb{N}$ nearest neighbors.
In later work \cite{bulyan}, the authors recognized that Krum may converge to an \emph{ineffectual} model in the landscape of non-convex high dimensional problems, such as in neural networks. 
They showed that a large adversarial change to a single parameter with a minor impact on the $L^p$ norm can make the model ineffective. In the same work, they present an alternative defense called \emph{Bulyan} to oppose such attacks. 
The algorithm works in two stages. In the first part, a \emph{selection set} of potentially benign values is iteratively constructed. In the second part, a variant of trimmed mean is applied to the selection set. 
Nevertheless, if $K$ machines are used, Bulyan is designed to defend only up to $(K-3)/4$ fraction of corrupted workers. 

Another category of defenses is based on \emph{redundancy} and seeks resilience to Byzantines by replicating the gradient computations such that each of them is processed by more than one machine \cite{lagrange_cdc, detox, draco, data_encoding}. Even though this approach requires more computation load, it comes with stronger guarantees of correcting the erroneous gradients. Existing redundancy-based techniques 
are sometimes combined with robust aggregation \cite{detox}. The main drawback of recent work in this category is that the training can be easily disrupted by a powerful, omniscient adversary that has full control of a subset of the nodes and can mount judicious attacks \cite{byzshield}.

Redundancy-based method \emph{DRACO} in \cite{draco} uses a simple \emph{Fractional Repetition Code} (FRC) (that operates by grouping workers) and the cyclic repetition code introduced in \cite{dimakis_cyclic_mds, tandon_gradient} to ensure robustness; majority voting and Fourier decoders try to alleviate the adversarial effects. Their work ensures exact recovery (as if the system had no adversaries) with $q$ Byzantine nodes, when each task is replicated $r \geq 2q+1$ times; the bound is information-theoretic minimum, and DRACO is not applicable if it is violated. Nonetheless, this requirement is very restrictive for the typical assumption that up to half of the workers can be Byzantine. 

\emph{DETOX} in \cite{detox} extends DRACO and uses a simple grouping strategy to assign the gradients. It performs multiple stages of aggregation to gradually filter the adversarial values. The first stage involves majority voting, while the following stages perform robust aggregation.
Unlike DRACO, the authors do not seek exact recovery; hence the minimum requirement in $r$ is small. However, the theoretical resilience guarantees that DETOX provides depend heavily on a ``random choice'' of the adversarial workers. In fact, we have crafted simple attacks \cite{byzshield} to make this aggregator fail under a more careful choice of adversaries. Furthermore, their theoretical results hold when the fraction of Byzantines is less than $1/40$. 

A third category focuses on \emph{ranking} and/or \emph{detection} \cite{regatti2020bygars, draco, zeno}; the objective is to rank workers using a reputation score to identify suspicious machines and exclude them or give them lower weight in the model update. This is achieved by means of computing reputation scores for each machine or by using ideas from coding theory to assign tasks to workers (encoding) and to detect the adversaries (decoding). \emph{Zeno} in \cite{zeno} ranks each worker using a score that depends on the estimated loss and the magnitude of the update. Zeno requires strict assumptions on the smoothness of the loss function and the gradient estimates' variance to tolerate an adversarial majority in the cluster. Similarly, \emph{ByGARS} \cite{regatti2020bygars} computes reputation scores for the nodes based on an auxiliary data set; these scores are used to weigh the contribution of each gradient to the model update. 


\subsection{Contributions}
In this paper, we propose novel techniques which combine \emph{redundancy}, \emph{detection}, and \emph{robust aggregation} for Byzantine resilience under a range of attack models and assumptions on the dataset/loss function. 

Our first scheme \emph{Aspis} is a subset-based assignment method for allocating tasks to workers in strong adversarial settings:  up to $q$ omniscient, colluding adversaries that can change at each iteration. We also consider weaker attacks: adversaries chosen randomly with limited collusion abilities, changing only after a few iterations at a time. It is conceivable that Aspis should continue to perform well with weaker attacks. However, as discussed later (Section \ref{sec:motivation_aspis_plus}), Aspis requires large batch sizes (for the mini-batch SGD). It is well-recognized that large batch sizes often cause performance degradation in training \cite{bottou_optimization}. Accordingly, for this class of attacks, we present a different algorithm called \emph{Aspis+} that can work with much smaller batch sizes. Both Aspis and Aspis+ use combinatorial ideas to assign the tasks to the worker nodes. Our work builds on our initial work in \cite{aspis_isit} and makes the following contributions.

\begin{itemize}
    \item 
    We demonstrate a worst-case upper bound (under any possible attack) on the fraction of corrupted gradients when Aspis is used. Even in this adverse scenario, our method enjoys a reduction in the fraction of corrupted gradients of more than 90\% compared with DETOX \cite{detox}. A weaker variation of this attack is where the adversaries do not collude and act randomly. In this case, we demonstrate that the Aspis protocol allows for detecting all the adversaries. In both scenarios, we provide theoretical guarantees on the fraction of corrupted gradients.
    
    \item In the setting where the dataset is distributed i.i.d. and the loss function is strongly convex and other technical conditions hold, we demonstrate a proof of convergence for Aspis. We demonstrate numerical results on the linear regression problem in this part; these show the advantage of Aspis over competing methods such as DETOX.
    
	
	
	
	\item For weaker attacks (discussed above), our experimental results indicate that Aspis+ detects all adversaries within approximately 5 iterations.
	\item 
	We present top-1 classification accuracy experiments on the CIFAR-10 \cite{cifar10} data set for various gradient distortion attacks coupled with choice/behavior patterns of the adversarial nodes. 
	Under the most sophisticated distortion methods \cite{alie}, the performance gap between Aspis/Aspis+ and other state-of-the-art methods is substantial, e.g., for Aspis it is 43\% in the strong scenario ({\it cf.} Figure \ref{fig:top1_fig_94}), and for Aspis+ 19\% in the weak scenario ({\it cf.} Figure \ref{fig:top1_fig_111}). 
\end{itemize}

\begin{figure}
	\begin{center}
		\includegraphics[width=0.48\textwidth]{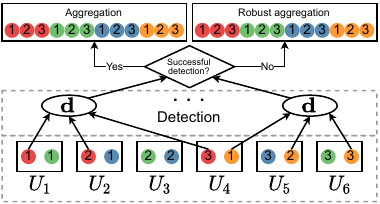}
	\end{center}
	\caption{Aggregation of gradients on a cluster.}
	\label{fig:aggregation_general}
\end{figure}

\section{Distributed Training Formulation}
\label{sec:formulation}
Assume a loss function $l_i(\mathbf{w})$ for the $i^\mathrm{th}$ sample of the dataset where $\mathbf{w}\in\mathbb{R}^d$ is the set of parameters of the model.\footnote{The paper's heavily-used notation is summarized in Appendix Table \ref{table:notation}.} The objective of distributed training is to minimize the empirical loss function $\hat{L}(\mathbf{w})$ with respect to $\bfw$, where
\begin{equation*}
	\hat{L}(\mathbf{w})=\frac{1}{n}\sum\limits_{i=1}^n l_i(\mathbf{w}).
\end{equation*}
Here $n$ denotes the number of samples.

We use either gradient descent (GD) or mini-batch Stochastic Gradient Descent (SGD) to solve this optimization. In both methods, initially $\mathbf{w}$ is randomly set to $\mathbf{w}_0$ ($\mathbf{w}_t$ is the model state at the end of iteration $t$). When using GD, the update equation is 
\begin{equation}
	\label{eq:gd_update}
	\mathbf{w}_{t+1}=\mathbf{w}_{t}-\eta_t \frac{1}{n}\sum\limits_{i=1}^n \nabla l_i(\mathbf{w}_t).
\end{equation}
Under mini-batch SGD a random \emph{batch} $B_t$ of $b$ samples is chosen to perform the update in the $t^{\mathrm{th}}$ iteration. Thus,
\begin{equation}
	\label{eq:vanilla_sgd_update}
	\mathbf{w}_{t+1}=\mathbf{w}_{t}-\eta_t\frac{1}{|B_t|}\sum\limits_{i\in B_t}\nabla l_i(\mathbf{w}_t).
\end{equation}
In both methods $\eta_t$ is the learning rate at the $t^{\mathrm{th}}$ iteration. The workers denoted $U_1, U_2, \dots, U_{K}$, compute gradients on subsets of the batch. The training is \emph{synchronous}, i.e., the PS waits for all workers to return before performing an update. It stores the data set and the model and coordinates the protocol. It can be observed that GD can be considered an instance of mini-batch SGD where the batch at each iteration is the entire dataset. Our discussion below is in the context of mini-batch SGD but can easily be applied to the GD case by using this observation.

We consider settings in this work that depend on the underlying assumptions on the dataset and the loss function. Setting-I does not make any assumption on the dataset or the loss function. In Setting-II at the top-level (technical details appear in Section \ref{sec:convergence_proof}) we assume that the data samples are distributed i.i.d. and the loss function is strongly-convex. The results that we provide depend on the underlying setting.

\textbf{Task assignment}: Each batch $B_t$ is split into $f$ disjoint subsets $\{B_{t,i}\}_{i=0}^{f-1}$, which are then assigned to the workers according to our placement policy. In what follows we refer to these as ``files'' to avoid confusion with other subsets that we need to refer to. Computational redundancy is introduced by assigning a given file to $r>1$ workers. As the load on all the workers is equal it follows that each worker is responsible for $l = fr/K$ files ($l$ is the \emph{computation load}).  We let $\calN^w(U_j)$ be the set of files assigned to worker $U_j$ and $\calN^f(B_{t,i})$ be the group of workers assigned to file $B_{t,i}$; our placement scheme is such that $\calN^f(B_{t,i})$ uniquely identifies the file $B_{t,i}$; thus, we will sometimes refer to the file $B_{t,i}$ by its worker assignment, $\calN^f(B_{t,i})$. We will also occasionally use the term \emph{group} (of the assigned workers) to refer to a file.  We discuss the actual placement algorithms used in this work in the upcoming subsection \ref{sec:task_assignment}.



\textbf{Training}: Each worker $U_j$ is given the task of computing the sum of the gradients on all its assigned files. For example, if file $B_{t,i}$ is assigned to $U_j$, then it calculates $\sum_{i' \in B_{t,i}} \nabla l_{i'}(\mathbf{w}_t)$ and returns them to the PS.
In every iteration, the PS will run our detection algorithm once it receives the results from all the users in an effort to identify the $q$ adversaries and will act according to the detection outcome.

Figure \ref{fig:aggregation_general} depicts this process. There are $K=6$ machines and $f=4$ distinct files (represented by colored circles) replicated $r=3$ times.\footnote{Some arrows and ellipses have been omitted from Figure \ref{fig:aggregation_general}; however, all files will be going through detection.} Each worker is assigned to $l=2$ files and computes the sum of gradients (or a distorted value) on each of them. The ``\textbf{d}'' ellipses refer to PS's detection operations immediately after receiving all the gradients.

{\bf Metrics}: We consider various metrics in our work. For Setting-I we consider (i) the fraction of distorted files, and (ii) the top-1 test accuracy of the final trained model. 
For the distortion fraction, let us denote the number of distorted files upon detection and aggregation by $c^{(q)}$ and its maximum value (under a worst-case attack) by $c_{\mathrm{max}}^{(q)}$. The \emph{distortion fraction} is $\epsilon := c^{(q)}/f$. The top-1 test accuracy is determined via numerical experiments. In Setting-II, in addition we consider proofs and rates of convergence of the proposed algorithms. We provide theoretical results and supporting experimental results on these.

\begin{table*}[!t]
	\centering
	\caption{Adversarial models considered in literature.}
	\label{table:adversarial_assumptions}
	\resizebox{1.5\columnwidth}{!}{
		\begin{tabular}{P{2cm}P{4cm}P{4.5cm}}
			\hline
			Scheme & Byzantine choice/orchestration & Gradient distortion \\
			\hline
			Draco \cite{draco} & optimal & reversed gradient, constant\\
			DETOX \cite{detox} & random & ALIE, reversed gradient, constant\\
			ByzShield \cite{byzshield} & optimal & ALIE, reversed gradient, constant\\
			Bulyan \cite{bulyan} & N/A & $\ell_2$-norm attack targeted on Bulyan\\
			Multi-Krum \cite{blanchard_krum} & N/A & random high-variance Gaussian vector\\
			Aspis & ATT-1, ATT-2 & ALIE, FoE, reversed gradient\\
			Aspis+ & ATT-3 & ALIE, constant\\
			\hline
		\end{tabular}
		}
\end{table*}

\subsection{Task Assignment}
\label{sec:task_assignment}
Let $\calU$ be the set of workers. Our scheme has $|\calU| \leq f$ (i.e., fewer workers than files). Our assignment of files to worker nodes is specified by a bipartite graph $\bfG_{task}$ where the left vertices correspond to the workers, and the right vertices correspond to the files. An edge in $\bfG_{task}$  between worker $U_i$ and a file $B_{t,j}$ indicates that the $U_i$ is responsible for processing file $B_{t,j}$.

\subsubsection{Aspis}
\label{sec:aspis_file_assignment}
For the Aspis scheme we construct $\bfG_{task}$ as follows. The left vertex set is $\{1, 2, \dots, K\}$ and the right vertex set corresponds to $r$-sized subsets of $\{1, 2, \dots, K\}$ (there are $\binom{K}{r}$ of them). An edge between $1 \leq i \leq K$ and $S \subset \{1, 2, \dots, K\}$ (where $|S| = r$) exists if $i \in S$. The worker set $\{U_1, \dots, U_K\}$ is in one-to-one correspondence with $\{1, 2, \dots, K\}$ and the files $B_{t,0}, \dots, B_{t, f-1}$ are in one-to-one correspondence with the $r$-sized subsets.

\begin{example}
	\label{ex:placement_K7_r3}
	Consider $K=7$ workers $U_1,U_2\dots,U_{7}$ and $r=3$. Based on our protocol, the $f=\binom{7}{3} =35$ files of each batch $B_t$ are associated one-to-one with 3-subsets of $\calU$, e.g., the subset $S = \{U_1,U_2,U_3\}$ corresponds to file $B_{t,0}$ and will be processed by $U_1$, $U_2$, and $U_3$.
\end{example}

\begin{remark}
	Our task assignment ensures that every pair of workers processes $\binom{K-2}{r-2}$ files. Moreover, the number of adversaries is $q < K/2$. Thus, upon receiving the gradients from the workers, the PS can examine them for consistency and flag certain nodes as adversarial if their computed gradients differ from $q+1$ or more of the other nodes. We use this intuition to detect and mitigate the adversarial effects and compute the fraction of corrupted files.
\end{remark}

\subsubsection{Aspis+}
\label{sec:aspis_plus_file_assignment}
For Aspis+, we use combinatorial designs \cite{DRSCDCA} to assign the gradient tasks to workers. Formally, a \textit{design} is a pair ($X$, $\mathcal{A}$) consisting of a set of $v$ elements (\textit{points}), $X$, and a family $\mathcal{A}$ (i.e., multiset) of nonempty subsets of $X$ called \textit{blocks}, where each block has the same cardinality $k$. Similar to Aspis, the workers and files are in one-to-one correspondence with the points and the blocks, respectively. Hence, for our purposes, the $k$ parameter of the design is the redundancy. A $t-(v,k,\lambda)$ design is one where any subset of $t$ points appear together in exactly $\lambda$ blocks. The case of $t=2$ has been studied extensively in the literature and is referred to as a \emph{balanced incomplete block design} (BIBD). A bipartite graph representing the incidence between the points and the blocks can be obtained naturally by letting the points correspond to the left vertices, and the blocks correspond to the right vertices. An edge exists between a point and a block if the point is contained in the block.

\begin{example}
	\label{ex:fano_plane}
	A $2-(7,3,1)$ design, also known as the \emph{Fano plane}, consists of the $v=7$ points $X = \{1,2,\dots,7\}$ and the block multiset $\mathcal{A}$ contains the blocks $\{1,2,3\}$, $\{1,4,7\}$, $\{2,4,6\}$, $\{3,4,5\}$, $\{2,5,7\}$, $\{1,5,6\}$ and $\{3,6,7\}$ with each block being of size $k=3$. In the bipartite graph $\bfG_{task}$ representation, we would have an edge, e.g., between point $2$ and blocks  $\{1,2,3\}, \{2,4,6\}$, and $\{2,5,7\}$.
\end{example}
In Aspis+ we construct $\bfG_{task}$ by the bipartite graph representing an appropriate $2-(v,k,\lambda)$ design. 



Another change compared to the Aspis placement scheme is that the points of the design will be randomly permuted at each iteration, i.e., for permutation $\pi$, the PS will map $\{U_1,U_2,\dots,U_K\} \xrightarrow{\pi} \{\pi(U_1),\pi(U_2),\dots,\pi(U_K)\}$. 
%
For instance, let us circularly permute the points of the Fano plane in Example \ref{ex:fano_plane} as $\pi(U_i) = U_{i+1}, i = 1,2,\dots,K-1$ and $\pi(U_K) = U_1$. Then, the file assignment at the next iteration will be based on the block collection $\mathcal{A} = \{\{2,3,4\},\{1,2,5\},\{3,5,7\},\{4,5,6\},\{1,3,6\},\{2,6,7\},\{1,\allowbreak4,7\}\}$. Permuting the assignment causes each Byzantine to disagree with more workers and to be detected in fewer iterations; details will be discussed in Section \ref{sec:aspis_plus_detection}. Owing to this permutation, we use a time subscript for the files assigned to $U_i$ for the $t^\mathrm{th}$ iteration; this is denoted by $\calN_t^w(U_i)$.

\section{Adversarial Attack Models and Gradient Distortion Methods}
\label{sec:all_attacks}

We now discuss the different Byzantine models that we consider in this work. For all the models, we assume that at most $q < K/2$ workers can be adversarial.
For each assigned file $B_{t,i}$ a worker $U_j$ will return the value $\hat{\mathbf{g}}_{t,i}^{(j)}$ to the PS. Then,
\begin{equation}
	\label{eq:returned_gradient}
	\hat{\mathbf{g}}_{t,i}^{(j)} = \left\{
	\begin{array}{ll}
		\mathbf{g}_{t,i} & \text{ if } U_j \text{ is honest},\\
		\mathbf{*} & \text{otherwise}, \\
	\end{array}
	\right.
\end{equation}
where $\mathbf{g}_{t,i}$ is the sum of the loss gradients on all samples in file $B_{t,i}$, i.e.,
\begin{equation*}
	\mathbf{g}_{t,i} = \sum\limits_{j\in B_{t,i}}\nabla l_j(\mathbf{w}_t)
\end{equation*}
and $\mathbf{*}$ is any arbitrary vector in $\mathbb{R}^d$.
Within this setup, we examine adversarial scenarios that differ based on the behavior of the workers. Table \ref{table:adversarial_assumptions} provides a high-level summary of the Byzantine models considered in this work as well as in related papers. As we will discuss in Section \ref{sec:experiments}, for those schemes that do not involve redundancy and merely split the work equally among the $K$ workers, all possible choices of the Byzantine set are equivalent, and no \emph{orchestration}\footnote{We will use the term \emph{orchestration} to refer to the method adversaries use to  collude and attack collectively as a group.} of them will change the defense's output; hence, those cases are denoted by ``N/A'' in the table.


\subsection{Attack 1}
\label{sec:attack_1_aspis}
We first consider a weak attack, denoted ATT-1, where the Byzantine nodes operate independently (i.e., do not collude) and attempt to distort the gradient on any file they participate in. For instance, a node may try to return arbitrary gradients on all its assigned files. For this attack, the identity of the workers may be arbitrary at each iteration as long as there are at most $q$ of them. 
\begin{remark}
	We emphasize that even though we call this attack ``weak'', this is the attack model considered in several prior works \cite{detox, draco}. To our best knowledge, most of them have not considered the adversarial problem from the lens of detection.
\end{remark}

\subsection{Attack 2}
\label{sec:attack_2_aspis}
Our second scenario, named ATT-2, is the strongest one we consider. We assume that the adversaries have full knowledge of the task assignment at each iteration and the detection strategies employed by the PS. The adversaries can collude in the ``best'' possible way to corrupt as many gradients as possible. Moreover, the set of adversaries can also change from iteration to iteration as long as there are at most $q$ of them.

\subsection{Attack 3}
\label{sec:attack_3_aspis_plus}
This attack is similar to ATT-1 and will be called ATT-3. On the one hand, it is weaker in the sense that the set of Byzantines (denoted $A$) does not change in every iteration. Instead, we will assume that there is a ``Byzantine window'' of $T_b$ iterations in which the set $A$ remains fixed. Also, the set $A$ will be a randomly chosen set of $q$ workers from $\calU$, i.e., it will not be chosen systematically. A new set will be chosen at random at all iterations $t$, where $t \equiv 0$ (mod $T_b$). Conversely, it is stronger than ATT-1 since we allow for limited collusion amongst the adversarial nodes. In particular, the Byzantines simulated by ATT-3 will distort only the files for which a Byzantine majority exists.

\subsection{Gradient Distortion Methods}
For each of the attacks considered above, the adversaries can distort the gradient in specific ways. Several such techniques have been considered in the literature and our numerical experiments use these methods for comparing different methods. For instance, \emph{ALIE} \cite{alie} involves communication among the Byzantines in which they jointly estimate the mean $\mu_i$ and standard deviation $\sigma_i$ of the batch's gradient for each dimension $i$ and subsequently use them to construct a distorted gradient that attempts to distort the median of the results. Another powerful attack is \emph{Fall of Empires (FoE)} \cite{FOE} which performs ``inner product manipulation'' to make the inner product between the true gradient and the robust estimator to be negative even when their distance is upper bounded by a small value. \emph{Reversed gradient} distortion returns $-c\*g$ for $c>0$, to the PS instead of the true gradient $\*g$. The \emph{constant attack} involves the Byzantine workers sending a constant gradient with all elements equal to a fixed value. To our best knowledge, the ALIE algorithm is the most sophisticated attack in literature for deep learning techniques. 

\section{Defense Strategies in Aspis and Aspis+}
\label{sec:detection}


In our work we use the Aspis task assignment and detection strategy for attacks ATT-1 and ATT-2. For ATT-3, we will use Aspis+. Recall that the methods differ in their corresponding task assignments. Nevertheless, the central idea in both detection methods is for the PS to apply a set of consistency checks on the obtained gradients from the different workers at each iteration to identify the adversaries. 

Let the current set of adversaries be $A \subset \{U_1,U_2,\dots,U_K\}$ with $|A|=q$; also, let $H$ be the honest worker set. The set $A$ is unknown, but our goal is to provide an estimate $\hat{A}$ of it. Ideally, the two sets should be identical. In general, depending on the adversarial behavior, we will be able to provide a set $\hat{A}$ such that $\hat{A} \subseteq A$. For each file, there is a group of $r$ workers which have processed it, and there are ${r\choose 2}$ pairs of workers in each group. Each such pair may or may not agree on the gradient value for the file. For iteration $t$, let us encode the agreement of workers $U_{j_1}, U_{j_2}$ on common file $i$ during the current iteration $t$ by 
\begin{equation}
	\alpha_{t,i}^{(j_1,j_2)} := \left\{
	\begin{array}{ll}
		1 & \text{if } \hat{\mathbf{g}}_{t,i}^{(j_1)} = \hat{\mathbf{g}}_{t,i}^{(j_2)},\\
		0 & \text{otherwise}.
	\end{array}
	\right.
\end{equation}

Across all files, the total number of agreements between a pair of workers $U_{j_1}, U_{j_2}$ during the $t^\mathrm{th}$ iteration is denoted by
\begin{equation}
	\alpha_t^{(j_1,j_2)} := \sum_{i\in \mathcal{N}_t^w(U_{j_1}) \cap \mathcal{N}_t^w(U_{j_2})}\alpha_{t,i}^{(j_1,j_2)}.
\end{equation}

Since the placement is known, the PS can always perform the above computation. Next, we form an undirected graph $\mathbf{G}_t$ whose vertices correspond to all workers $\{U_1, U_2, \dots, U_{K}\}$. An edge $(U_{j_1}, U_{j_2})$ exists in $\mathbf{G}_t$ only if the computed gradients (at iteration $t$) of $U_{j_1}$ and $U_{j_2}$ match in ``all'' their common assignments.


\subsection{Aspis Detection Rule}
\label{sec:aspis_detection}
In what follows, we suppress the iteration index $t$ since the Aspis algorithm is the same for each iteration.
For the Aspis task assignment ({\it cf.} Section \ref{sec:aspis_file_assignment}), any two workers, $U_{j_1}$ and $U_{j_2}$, have ${{K-2}\choose {r-2}}$ common files.

Let us index the $q$ adversaries in $A = \{A_1,A_2,\dots,A_q\}$ and the honest workers in $H$. We say that two workers $U_{j_1}$ and $U_{j_2}$ disagree if there is no edge between them in $\mathbf{G}$. The non-existence of an edge between $U_{j_1}$ and $U_{j_2}$ only means that they disagree in \emph{at least one} of the $\binom{K-2}{r-2}$ files that they jointly participate in. For corrupting the gradients, each adversary has to disagree on the computations with a subset of the honest workers. An adversary may also disagree with other adversaries.  

A \emph{clique} in an undirected graph is defined as a subset of vertices with an edge between any pair of them. A \emph{maximal clique} is one that cannot be enlarged by adding additional vertices to it. A \emph{maximum clique} is one such that there is no clique with more vertices in the given graph. We note that the set of honest workers $H$ will pair-wise agree on all common tasks. Thus, $H$ forms a clique (of size $K-q$) within $\mathbf{G}$. The clique containing the honest workers may not be maximal. However, it will have a size of at least $K-q$. Let the maximum clique on $\mathbf{G}$ be $M_{\mathbf{G}}$. Any worker $U_j$ with $\deg(U_j) < K-q-1$ will not belong to a maximum clique and can right away be eliminated as a ``detected'' adversary.

\begin{algorithm}[!t]
	\KwIn{Computed gradients $\hat{\mathbf{g}}_{t,i}^{(j)}$, $i=0,1,\dots,f-1$, $j=1,2,\dots,K$, redundancy $r$ and empty graph $\mathbf{G}$ with worker vertices $\calU$.}
	{
		\abovedisplayskip=0pt
		\belowdisplayskip=0pt
		\For{each pair $(U_{j_1}, U_{j_2}), j_1\neq j_2$ of workers}{
			PS computes the number of agreements $\alpha^{(j_1,j_2)}$ of the pair $U_{j_1}, U_{j_2}$ on the gradient value.
			
			\If{$\alpha^{(j_1,j_2)} = {{K-2}\choose {r-2}}$}{
				Connect vertex $U_{j_1}$ to vertex $U_{j_2}$ in $\mathbf{G}$.
			}
		}
		
		PS enumerates all $k$ maximum cliques $M_{\mathbf{G}}^{(1)}, M_{\mathbf{G}}^{(2)}, \dots, M_{\mathbf{G}}^{(k)}$ in $\mathbf{G}$.
		
		\eIf{there is a unique maximum clique $M_{\mathbf{G}}$ ($k=1$)}{
			PS determines the honest workers $H = M_{\mathbf{G}}$ and the adversarial machines $\hat{A} = \mathcal{U} - M_{\mathbf{G}}$.
		}{
			PS declares unsuccessful detection.
		}
	}
	\caption{Proposed Aspis graph-based detection.}
	\label{alg:detection}
\end{algorithm}

\begin{algorithm}[!t]
	\KwIn{
		Data set of $n$ samples, batch size $b$, computation load $l$, redundancy $r$, \newline number of files $f$, maximum iterations $T$, file assignments $\{\calN^w(U_i)\}_{i=1}^{K}$, robust estimator function $\widehat{\mathrm{med}}$.
	}
	{
		\abovedisplayskip=0pt
		\belowdisplayskip=0pt
		The PS randomly initializes model's parameters to $\mathbf{w}_0$.\\
		\For{$t = 0$ to $T-1$}{
			PS chooses a random batch $B_t\subseteq\{1,2,\dots,n\}$ of $b$ samples, partitions it into $f$ files $\{B_{t,i}\}_{i=0}^{f-1}$ and assigns them to workers according to $\{\calN^w(U_i)\}_{i=1}^{K}$. It then transmits $\mathbf{w}_t$ to all workers.\\
			\For{each worker $U_j$}{
				\eIf{$U_j$ is honest}{
					\For{each file $i \in \calN^w(U_j)$}{
						$U_j$ computes the sum of gradients $$\hat{\mathbf{g}}_{t,i}^{(j)}=\sum\limits_{k\in B_{t,i}}\nabla l_k(\mathbf{w}_t).$$
					}
				}{
					$U_j$ constructs $l$ adversarial vectors $$\hat{\mathbf{g}}_{t,i_1}^{(j)},\hat{\mathbf{g}}_{t,i_2}^{(j)},\dots,\hat{\mathbf{g}}_{t,i_l}^{(j)}.$$
				}
				$U_j$ returns $\hat{\mathbf{g}}_{t,i_1}^{(j)},\hat{\mathbf{g}}_{t,i_2}^{(j)},\dots,\hat{\mathbf{g}}_{t,i_l}^{(j)}$ to the PS.
			}
			PS runs a detection algorithm to identify the adversaries.
			
			\eIf{detection is successful}{
				Let $H$ be the detected honest workers. Initialize a non-corrupted gradient set as $\calG = \emptyset$.\\
				\For{each file in $\{B_{t,i}\}_{i=0}^{f-1}$}{
					PS chooses the gradient of a worker in $\calN^f(B_{t,i}) \cap H$ (if non-empty) and adds it to $\calG$.
				}
				\begin{equation*}
					\mathbf{w}_{t+1}=\mathbf{w}_{t}-\eta_t\frac{1}{|\calG|}\sum\limits_{\mathbf{g}\in \calG}\mathbf{g}.
				\end{equation*}
			}{
				\For{each file in $\{B_{t,i}\}_{i=0}^{f-1}$}{
					PS determines the $r$ workers in $\calN^f(B_{t,i})$ which have processed $B_{t,i}$ and computes $$\mathbf{m}_i = \mathrm{majority}\left\{\hat{\mathbf{g}}_{t,i}^{(j)}: U_j \in \calN^f(B_{t,i})\}\right\}.$$
				}
				PS updates the model via
				\begin{equation*}
					\mathbf{w}_{t+1} = \mathbf{w}_{t} - \eta_t \times \widehat{\mathrm{med}}\{\mathbf{m}_i: i = 0,1,\dots,f-1\}.
				\end{equation*}
			}
		}
	}
	\caption{Proposed Aspis/Aspis+ aggregation protocol to alleviate Byzantine effects.}
	\label{alg:main_algorithm}
\end{algorithm}

The essential idea of our detection is to run a \emph{clique-finding} algorithm on $\mathbf{G}$ (summarized in Algorithm \ref{alg:detection}). The detection may be successful or unsuccessful depending on which attack is used; we discuss this in more detail shortly.


We note that clique-finding is well-known to be an NP-complete problem \cite{karp1972}. Nevertheless, there are fast, practical algorithms with excellent performance on graphs even up to hundreds of nodes \cite{cazals_clique, tomita_clique}. Specifically, the authors of \cite{tomita_clique} have shown that their proposed algorithm, which enumerates all maximal cliques, has similar complexity as other methods \cite{robson_1986, tarjan_1977}, which are used to find a single maximum clique. We utilize this algorithm. 
Our extensive experimental evidence suggests that clique-finding is not a computation bottleneck for the size and structure of the graphs that Aspis uses. We have experimented with clique-finding on a graph of $K=100$ workers and $r=5$ for different values of $q$; in all cases, enumerating all maximal cliques took no more than 15 milliseconds. These experiments and the asymptotic complexity of the entire protocol are addressed in Supplement Section \ref{appendix:asymptotic}. 
During aggregation (see Algorithm \ref{alg:main_algorithm}), the PS will perform a majority vote across the computations of each file (implementation details in Supplement Section \ref{appendix:gradient_equality}). Recall that $r$ workers have processed each file. For each such file $B_{t,i}$, the PS decides a majority value $\mathbf{m}_i$

\begin{equation}
	\label{eq:basic_majority}
	\mathbf{m}_i := \mathrm{majority}\left\{\hat{\mathbf{g}}_{t,i}^{(j)}: U_j \in \calN^f(B_{t,i})\right\}.
\end{equation}


Assume that $r$ is odd and let $r'=\frac{r+1}{2}$. Under the rule in Eq. \eqref{eq:basic_majority}, the gradient on a file is distorted only if at least $r'$ of the computations are performed by Byzantines. Following the majority vote, we will further filter the gradients using a robust estimator $\widehat{\mathrm{med}}$ (see Algorithm \ref{alg:main_algorithm}, line 25). This robust estimator is either the coordinate-wise median or the geometric median; a similar setup was considered in \cite{byzshield, detox}. For example, in Figure \ref{fig:aggregation_general}, all returned values for the red file will be evaluated by a majority vote function on the PS, which decides a single output value; a similar voting is done for the other 3 files. After the voting process, Aspis applies the robust estimator $\widehat{\mathrm{med}}$ on the ``winning'' gradients $\mathbf{m}_i$, $i=0,1,\dots,f-1$.


\begin{figure}[t]
	\centering
	\begin{subfigure}[b]{0.43\textwidth}
		\centering
		\includegraphics[scale=0.3]{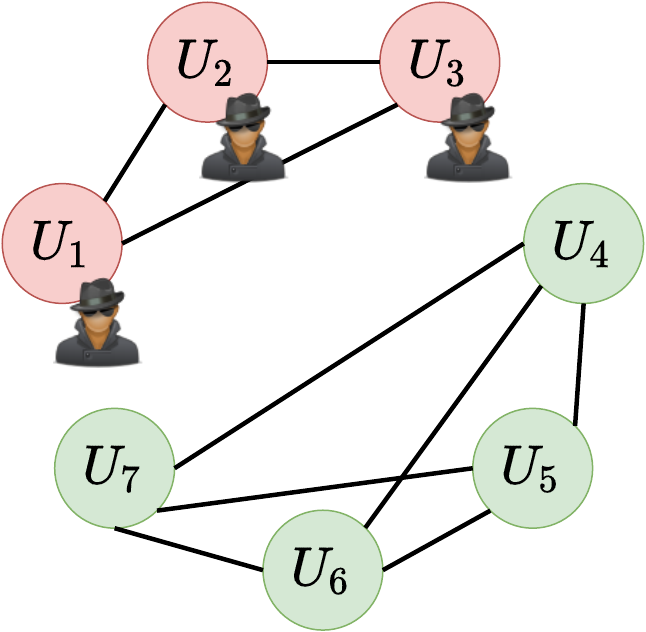}
		\caption{Unique max-clique, detection succeeds.}
		\label{fig:subset_assignment_K7_r3_graph_success}
	\end{subfigure}
	\hspace{0.01\textwidth}
	\begin{subfigure}[b]{0.43\textwidth}
		\centering
		\includegraphics[scale=0.3]{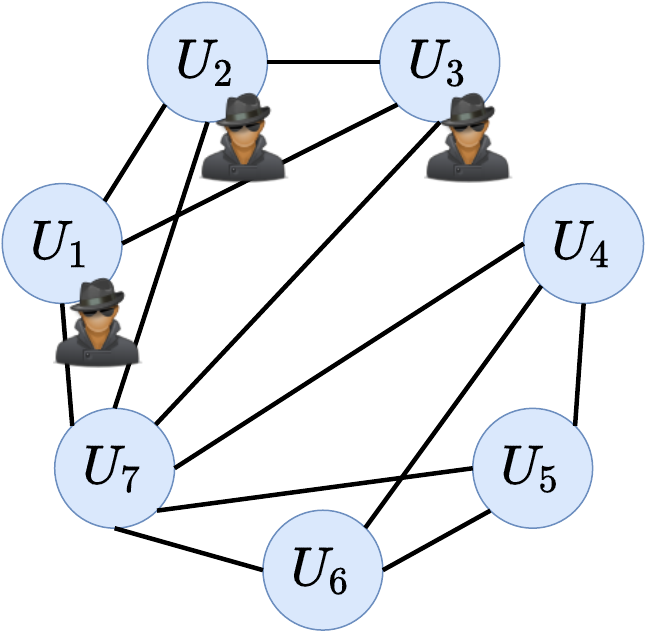}
		\caption{Two max-cliques, detection fails.}
		\label{fig:subset_assignment_K7_r3_graph_failure}
	\end{subfigure}
	\caption{Detection graph $\mathbf{G}$ for $K=7$ workers among which $U_1$, $U_2$ and $U_3$ are the adversaries.}
\end{figure}

\subsubsection{Defense Strategy Against ATT-1}
\label{sec:weak_subset_attack}
Under ATT-1, it is clear that a Byzantine node will disagree with at least $K-q$ honest nodes (as, by assumption in Section \ref{sec:attack_1_aspis}, it will disagree with all of them), and thus, the degree of the node in $\mathbf{G}$ will be at most $q-1 < K-q-1$, and it will not be part of the maximum clique. Thus, each of the adversaries will be detected, and their returned gradients will not be considered further. The algorithm declares the (unique) maximum clique as honest and proceeds to aggregation. In particular, assume that $h$ workers $U_{i_1}, U_{i_2}, \dots, U_{i_h}$ have been identified as honest. For each of the $f$ files, if at least one honest worker processed it, the PS will pick one of the ``honest'' gradient values. The chosen gradients are then averaged for the update (\emph{cf.} Eq. \eqref{eq:vanilla_sgd_update}). For instance, in Figure \ref{fig:aggregation_general}, assume that $U_1$, $U_2$, and $U_4$ have been identified as faulty. During aggregation, the PS will ignore the red file as all 3 copies have been compromised. For the orange file, it will pick either the gradient computed by $U_5$ or $U_6$ as both of them are ``honest.'' The only files that can be distorted in this case are those that consist exclusively of adversarial nodes.

Figure \ref{fig:subset_assignment_K7_r3_graph_success} (corresponding to Example \ref{ex:placement_K7_r3}) shows an example where in a cluster of size $K=7$, the $q=3$ adversaries are $A = \{U_1, U_2, U_3\}$ and the remaining workers are honest with $H = \{U_4, U_5, U_6, U_7\}$. In this case, the unique maximum clique is $M_{\mathbf{G}} = H$, and detection is successful. Under this attack, the distorted files are those whose all copies have been compromised, i.e., $c^{(q)} = \binom{q}{r}$.


\subsubsection{Defense Strategy Against ATT-2 (Robust Aggregation)}
\label{sec:optimal_subset_attack} 
Let $D_i$ denote the set of disagreement workers for adversary $A_i, i = 1,2,\dots, q$, where $D_i$ can contain members from $A$ and from $H$. If the attack ATT-2 is used on Aspis, upon the formation of $\mathbf{G}$ we know that a worker $U_j$ will be flagged as adversarial if $deg(U_j) < K - q -1$. Therefore to avoid detection, a \emph{necessary} condition is that $|D_j|\leq q$. 

We now upper bound the number of files that can be corrupted under {\it any possible strategy} employed by the adversaries. Note that according to Algorithm \ref{alg:main_algorithm}, we resort to robust aggregation in case of more than one maximum clique in $\mathbf{G}$. In this scenario, a gradient can only be corrupted if a majority of the assigned workers computing it are adversarial and agree on a wrong value. The proof of the following theorem appears in Appendix Section \ref{appendix:fixed_diagreement_optimality}.
\begin{theorem}
	\label{theorem:aspis_optimal_attack}
	Consider a training cluster of $K$ workers with $q$ adversaries using algorithm in Section \ref{sec:aspis_file_assignment} to assign the $f = \binom{K}{r}$ files to workers, and Algorithm \ref{alg:detection} for adversary detection. Under any adversarial strategy, the maximum number of files that can be corrupted is 
	\begin{equation}
		c_{\mathrm{max}}^{(q)} = \frac{1}{2}{2q\choose r}.
	\end{equation}
	Furthermore, this upper bound can be achieved if all adversaries fix a set $D \subset H$ of honest workers with which they will consistently disagree on the gradient (by distorting it).
\end{theorem}

\begin{remark}
    We emphasize that the maximum fraction of corrupted gradients $c_{\mathrm{max}}^{(q)}/f$ is much lesser as compared to the baseline $q/K$ and with respect to other schemes as well (details in Sec. \ref{sec:distortion_fraction_analysis}). For instance  $K=15$ and $q=3$, at most 0.022 fraction of the gradients are corrupted in Aspis as against 0.2 for the baseline scheme.
\end{remark}
In Appendix Section \ref{appendix:fixed_diagreement_optimality} we show that under ATT-2 there is bound to be more than one maximum clique in the detection graph. Thus, the PS cannot unambiguously decide which one is the honest one; detection fails and we fall back to the robust aggregation technique.

An example is shown in Figure \ref{fig:subset_assignment_K7_r3_graph_failure} for the setup of Example \ref{ex:placement_K7_r3}. The adversaries $A = \{U_1, U_2, U_3\}$ consistently disagree with the workers in $D = \{U_4, U_5, U_6\} \subset H$. The ambiguity as to which of the two maximum cliques ($\{U_1, U_2, U_3, U_7\}$ or $\{U_4, U_5, U_6, U_7\}$) is the honest one makes an accurate detection impossible; robust aggregation will be performed instead.

\subsection{Motivation for Aspis+}
\label{sec:motivation_aspis_plus}
Our motivation for proposing Aspis+ originates in the limitations of the subset assignment of Aspis. It is evident from the experimental results in Section \ref{sec:experiments} that Aspis is more suitable to worst-case attacks where the adversaries collude and distort the maximum number of tasks in an undetected fashion; in this case, the accuracy gap between Aspis and prior methods is maximal. Aspis does not perform as well under weaker attacks such as the \emph{reversed gradient} attack (\emph{cf.} Figures \ref{fig:top1_fig_97}, \ref{fig:top1_fig_98}, \ref{fig:top1_fig_99} even though it achieves a much smaller distortion fraction $\epsilon$, as discussed in Section \ref{sec:distortion_fraction_analysis}. This can be attributed to the fact that the number of tasks is $K\choose r$ and even for the considered cluster of $K=15$, $r=3$, it would require splitting the batch into 455 files; hence, the batch size must be a multiple of 455. There is significant evidence that large batch sizes can hurt generalization and make the model converge slowly \cite{bottou_optimization, masters_small_batch_training, you_large_batch_training}. Some workarounds have been proposed to solve this problem. For instance, the work of \cite{you_large_batch_training} uses layer-wise adaptive rate scaling to update each layer using a different learning rate. The authors of \cite{geiping_stochastic_training} perform implicit regularization using warmup and cosine annealing to tune the learning rate as well as gradient clipping. However, these methods require training for a significantly larger number of epochs. For the above reasons, we have extended our work and proposed Aspis+ to handle weaker Byzantine failures ({\it cf.} ATT-3) while requiring a much smaller batch size. 

\begin{algorithm}[!t]
	\KwIn{Computed gradients $\hat{\mathbf{g}}_{t,i}^{(j)}$, $i=0,1,\dots,f-1$, $j=1,2,\dots,K$, $2-(v,k,\lambda)$ design, length of detection window $T_d$, maximum iterations $T$.}
	{
		\abovedisplayskip=0pt
		\belowdisplayskip=0pt
		\For{$t = 0$ to $T-1$}{
			Let $t' = t\ (\mathrm{mod}\ T_d) + 1$.\\
			\If{$t' = 1$}{
				Set $\mathbf{G}$ as the complete graph with worker vertices $\calU$.\\
				$\forall j_1, j_2$, set  $\alpha^{(j_1,j_2)} = 0$.
			}
			\For{each pair $(U_{j_1}, U_{j_2}), j_1\neq j_2$ of workers}{
				PS computes the number of agreements $\alpha_t^{(j_1,j_2)}$ of the pair $U_{j_1}, U_{j_2}$ on the gradient value.
				
				Update $\alpha^{(j_1,j_2)} = \alpha^{(j_1,j_2)} + \alpha_t^{(j_1,j_2)}$.
			}
			\For{each pair $(U_{j_1}, U_{j_2}), j_1\neq j_2$ of workers}{			
				\If{$\alpha^{(j_1,j_2)} < \lambda\times t'$}{
					Remove edge $(U_{j_1}, U_{j_2})$ from $\mathbf{G}$.
				}
			}
			\For{each worker $U_j \in \calU$}{			
				\If{$deg(U_j) < K - q - 1$}{
					$\hat{A} = \hat{A} \cup \{U_j\}$.
				}
			}
			\If{$|\hat{A}| > q$}{
				Set $\hat{A}$ to be the $q$ most recently detected Byzantines.
			}
		}

	}
	\caption{Proposed Aspis+ graph-based detection.}
	\label{alg:detection_aspis_plus}
\end{algorithm}

\subsection{Aspis+ Detection Rule}
\label{sec:aspis_plus_detection}

The principal intuition of the Aspis+ detection approach (used for ATT-3) is to iteratively keep refining the graph $\bfG$ in which the edges encode the agreements of workers during consecutive and non-overlapping windows of $T_d$ iterations. 
At the beginning of each such window, the PS will reset $\mathbf{G}$ to be a complete graph, i.e., as if all workers pairwise agree with other. Then, it will gradually remove edges from $\mathbf{G}$ as disagreements between the workers are observed; hence, the graph will be updated at each of the $T_d$ iterations of the window, and the PS will assume that the Byzantine set does not change within a detection window. In practice, as we do not know the ``Byzantine window,'' we will not assume an alignment between the two kinds of windows, and we will set $T_d \neq T_b$ for our experiments. The detection method will be the same for all detection windows; thus, we will analyze the process in one window of $T_d$ steps. 

For a detection window, let us encode the agreement of workers $U_{j_1}, U_{j_2}$ on common file $i$ during the current iteration $t$ of the window $t = 1,2,\dots,T_d$ as
\begin{equation}
	\alpha_{t,i}^{(j_1,j_2)} := \left\{
	\begin{array}{ll}
		1 & \text{if } \hat{\mathbf{g}}_{t,i}^{(j_1)} = \hat{\mathbf{g}}_{t,i}^{(j_2)},\\
		0 & \text{otherwise}.
	\end{array}
	\right.
\end{equation}
Across all files, the total number of agreements between a pair of workers $U_{j_1}, U_{j_2}$ during the $t^\mathrm{th}$ iteration is denoted by
\begin{equation}
	\alpha_t^{(j_1,j_2)} := \sum_{i\in \mathcal{N}_t^w(U_{j_1}) \cap \mathcal{N}_t^w(U_{j_2})}\alpha_{t,i}^{(j_1,j_2)}.
\end{equation}
Assume that the current iteration of the window is indexed with $t'\in\{1,2,\dots,T_d\}$. The PS will collect all agreements for each pair of workers $U_{j_1}, U_{j_2}$ up until the current iteration as
\begin{equation}
	\label{eq:alpha_workers_window}
	\alpha^{(j_1,j_2)} := \sum\limits_{t=1}^{t'} \alpha_t^{(j_1,j_2)}.
\end{equation}
Since the placement is known, the PS can always perform the above computation. Next, it will examine the agreements and update $\mathbf{G}$ as necessary. 

Based on the task placement (\emph{cf.} Section \ref{sec:aspis_plus_file_assignment}), an edge $(U_{j_1}, U_{j_2})$ exists in $\mathbf{G}$ only if the computed gradients of $U_{j_1}$ and $U_{j_2}$ match in all their $\lambda$ common groups in all iterations up to the current one indexed with $t'$, i.e., a pair $U_{j_1}, U_{j_2}$ needs to have $\alpha^{(j_1,j_2)} = \lambda\times t'$ for an edge $(U_{j_1}, U_{j_2})$ to be in $\mathbf{G}$. If this is not the case, the edge $(U_{j_1}, U_{j_2})$ will be removed from $\mathbf{G}$. After all such edges are examined, detection is done using degree counting. Given that there are $q$ Byzantines in the cluster, after examining all pairs of workers and determining the form of $\mathbf{G}$, a worker $U_j$ will be flagged as Byzantine if $deg(U_j) < K - q - 1$. Based on Eq. \eqref{eq:alpha_workers_window}, it is not hard to see that such workers can be eliminated and their gradients will not be considered again until the last iteration of the current window. The only exception to this is if the Byzantine set changes before the end of the current detection window. This is possible due to a potential misalignment between the ``Byzantine window'' and the detection window (recall that $T_d \neq T_b$ is assumed to avoid trivialities). In this case, more than $q$ workers may be detected as Byzantines; the PS will, by convention, choose $\hat{A}$ to be the most recently detected Byzantines. 
Algorithm \ref{alg:detection_aspis_plus} discusses the detection protocol. Following detection, the PS will act as follows. 
If at least one Byzantine has been detected, it will ignore the votes of detected Byzantines, and for each group, if there is at least one ``honest'' vote, it will use this as the output of the majority voting group; also, if a group consists merely of detected Byzantines, it will completely ignore the group. The remaining groups will go through robust aggregation (as in Section \ref{sec:aspis_detection}). In our experiments in Section \ref{sec:experiments_aspis+}, all Byzantines are detected successfully in at most 5 iterations. Example \ref{ex:permutation_detection_example} showcases the utility of permutations in our detection algorithm using $K=7$ workers.

\begin{example}
	\makeatletter
	\def\old@comma{,}
	\catcode`\,=13
	\def,{%
		\ifmmode%
		\old@comma\discretionary{}{}{}%
		\else%
		\old@comma%
		\fi%
	}
	\makeatother
	
	\label{ex:permutation_detection_example}
	We will use the assignment of Example \ref{ex:fano_plane} with $K=7$ workers $\calU = \{1,2,\dots,7\}$ assigned to tasks according to a $2-(7,3,1)$ Fano plane and let us denote the assignment of workers to groups (blocks of the design) during the $t^\mathrm{th}$ iteration by $\mathcal{A}_t$, initially equal to $\mathcal{A}_1 = \{\{1,2,3\}, \{1,4,7\}, \{2,4,6\}, \{3,4,5\}, \{2,5,7\}, \{1,5,6\}, \{3,6,7\}\}$. For the windows, assume that $T_d>2$ and $T_b>2$. Also, let $q = 2$ and the Byzantine set be $A = \{U_1,U_2\}$. Based on ATT-3, workers $U_1,U_2$ are in majority within a group in which they disagree with worker $U_3$. After the first permutation, a possible assignment is $\mathcal{A}_2 = \{\{1,3,6\}, \{3,4,7\}, \{2,4,6\}, \{1,4,5\}, \{5,6,7\}, \{2,3,5\}, \{1,2,7\}\}$. Then, $U_1,U_2$ are in the same group as the honest $U_7$ with which they disagree; hence, $deg(U_1) = deg(U_2) = 4 = K-q-1$, and none of them affords to disagree with more honest workers to remain undetected. However, if the next permutation assigns the workers as $\mathcal{A}_3 = \{\{1,3,6\}, \{1,4,7\}, \{4,6,5\}, \{2,3,4\}, \{2,6,7\}, \{1,2,5\}, \{3,5,7\}\}$ then the adversaries will cast a different vote than $U_5$ as well. Both of them will be detected after only three iterations.
\end{example}

\begin{remark}
	Using a $2-(v,3,\lambda)$, i.e., a design with $k=3$ (a typical value for the redundancy) to assign the files on a cluster with $q$ Byzantines, the maximum number of files one can distort is $\lambda{q\choose2}/|\calB|$ \cite{DRSCDCA}, where $|\calB|$ is the total number of files; this is when each possible pair of Byzantines, among the $q\choose2$ possible ones appear together in a distinct block and distorts the corresponding file. In Aspis+, the focus is on weak attacks and determining the worst-case choice of adversaries that maximize the number of distorted files is beyond the scope of our work.
\end{remark}

\begin{figure}[t]
	\centering
	\begin{subfigure}[b]{0.43\textwidth}
		\centering
		\includegraphics[scale=0.4]{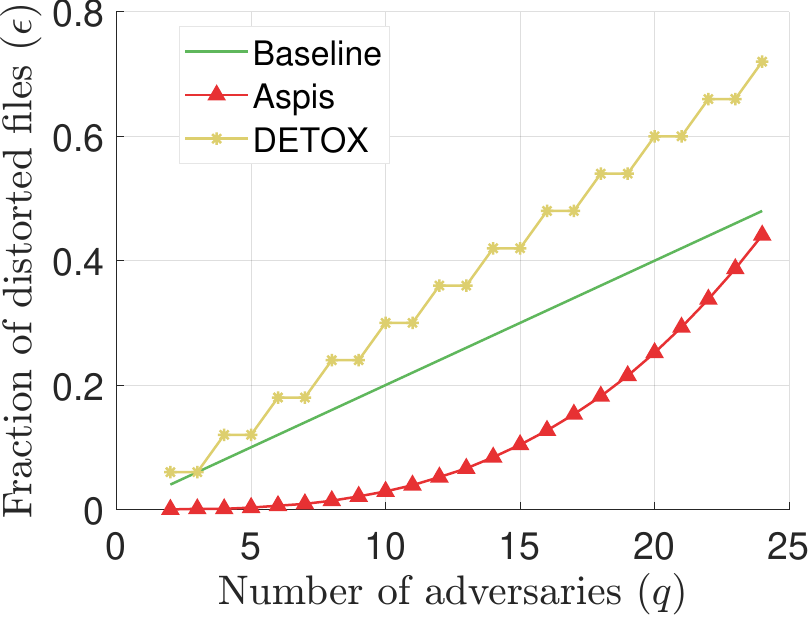}
		\caption{Optimal attacks.}
		\label{fig:epsilon_simulations_optimal_K50}
	\end{subfigure}
	\hspace{0.01\textwidth}
	\begin{subfigure}[b]{0.43\textwidth}
		\centering
		\includegraphics[scale=0.4]{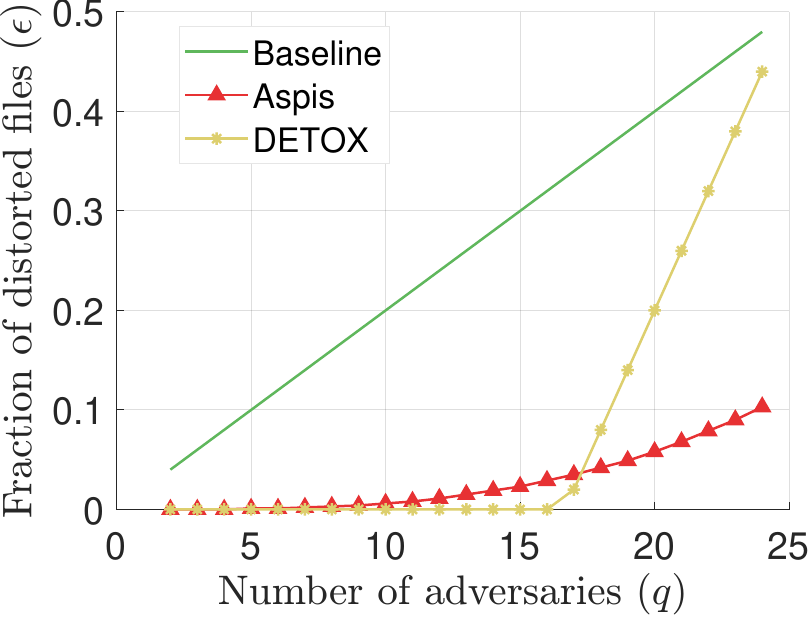}
		\caption{Weak attacks.}
		\label{fig:epsilon_simulations_weak_K50}
	\end{subfigure}
	\caption{Distortion fraction of optimal and weak attacks for $(K,r)=(50,3)$ and comparison.}
\end{figure}

\section{Convergence Results and Experiments under Setting-II}
\label{sec:convergence_proof}
In this section, we operate under Setting-II ({\it cf.} Section \ref{sec:formulation}). By leveraging the work of Chen et al. \cite{ChenSX17} we demonstrate that our training algorithm converges to the optimal point. 

We assume that the data samples are distributed i.i.d. from some unknown distribution $\mu$. We are interested in finding $\bfw^*$ that minimizes $L(\bfw) = \mathbb{E}(l_1(\bfw))$ over the $\bfw \in \mathcal{W}$; here the expectation is over the distribution $\mu$ and the $l_i(\bfw)$'s are distributed i.i.d as well. In general, since the distribution is unknown, $\mathbb{E}(l_1(\bfw))$ cannot be computed and we instead minimize the empirical loss function given by $\hat{L}(\bfw) = \frac{1}{n} \sum_{i=1}^n l_i(\bfw)$. We need the following additional assumptions.

In the discussion below, we say that a random vector $\bfz$ is sub-exponential with sub-exponential norm $K$ if for every unit-vector $v$, $\bfz^\top v$ is a sub-exponential random variable with sub-exponential norm at most $K$, i.e., $ \sup_{v: ||v|| \le 1} Pr(|\bfz^\top v|  > t ) \leq \exp(- t/ K)$ \cite[Sec 2.7]{versh_book}. To keep notation simple, we reuse the letter $C$ to denote different numerical constants in each use. This practice is common when working with classes of distributions such as sub-exponential.

\begin{itemize}
    
    \item The minimization of $\hat{L}(\bfw)$ is performed by using Aspis or Aspis+ along with gradient descent ({\it cf.} Eq. \eqref{eq:gd_update}). This means that in Algorithm \ref{alg:main_algorithm}, the batch size $b=n$ for all iterations ({\it cf.} discussion after \eqref{eq:vanilla_sgd_update}). The robust estimator $\widehat{\mathrm{med}}$ is the geometric median.
    
    \item The function $L(\bfw)$ is $\beta-$strongly convex, and differentiable with respect to $\bfw$ with $\tilde{M}$-Lipschitz gradient. This means that for all $\bfw$ and $\bfw'$ we have
    \begin{align*}
        &L(\bfw') \geq L(\bfw) + \nabla L(\bfw)^T(\bfw'- \bfw) + \frac{\beta}{2} \norm{\bfw - \bfw'}^2, \text{~and}\\
        &\norm{\nabla L(\bfw) - \nabla L(\bfw')} \leq \tilde{M} \norm{\bfw - \bfw'}.
    \end{align*}
    
    \item The random vectors $\nabla l_i(\bfw)$ for $i=1,\dots, n$ are sub-exponential with sub-exponential norm $C$.
    This assumption ensures that $\frac{1}{n} \sum_{i=1}^n \nabla l_i(\bfw^*)$ concentrates around its mean $\nabla L(\bfw^*) = 0$.
    
    \item Let $h_i(\bfw) = \nabla l_i(\bfw) - \nabla l_i(\bfw^*)$. For $i=1, \dots, n$, the random vectors $h_i(\bfw)$ are sub-exponential with sub-exponential norm $C\norm{\bfw - \bfw^*}$. 
    \item For any $\delta \in (0,1)$ there exists $\tilde{M}'$ (dependent on $n$ and $\delta$) that is non-increasing in $n$ such that $\hat{L}(\bfw)$ is $\tilde{M}'$-smooth with high probability, i.e,
    \begin{align*}
        & P \left( \sup_{\bfw, \bfw' \in \calW: \bfw \neq \bfw'} \frac{\norm{\frac{1}{n} \sum_{i=1}^n (\nabla l_i(\bfw) - \nabla l_i(\bfw')) }}{\norm{\bfw - \bfw'}} \leq \tilde{M}'\right)\\
        & \geq 1 - \frac{\delta}{3}.
    \end{align*}
    Here $\calW$ is the feasible parameter set.
\end{itemize}

For Aspis, Theorem \ref{theorem:aspis_optimal_attack} guarantees an upper bound on the fraction of corrupted gradients regardless of what attack is used. 
In particular, treating the majority logic and clique finding as a pre-processing step, we arrive at a set of $f$ files, at most $c^{(q)}_{\max}$ ({\it cf.} Theorem \ref{theorem:aspis_optimal_attack}) of which are ``arbitrarily'' corrupted. At this point, the PS applies the robust estimator $\widehat{\mathrm{med}}$ - ``geometric median'' and uses it to perform the update step. We can leverage Theorem 5 of \cite{ChenSX17} to obtain the following result where $d$ is the length of the parameter vector and for $p_i \in (0,1), i = 1,2$ the quantity $D(p_1 || p_2) = p_1 \log_2 (\frac{p_1}{p_2}) + (1-p_1) \log_2 (\frac{1-p_1}{1-p_2})$.
\begin{theorem} (adapted from \cite{ChenSX17})
Suppose that $\beta,\tilde{M}$ are all constants and $\log \tilde{M}' = \mathcal{O}(\log d)$. Assume that $\calW \subset \{\bfw ~:~ \norm{\bfw - \bfw^*} \leq \tilde{r} \sqrt{d}\}$ for positive $\tilde{r}$ such that $\log \tilde{r} =  \mathcal{O}(d \log (n/f))$ and $2(1+\epsilon)  c^{(q)}_{\max} \leq f$. Fix any $\alpha \in (c^{(q)}_{\max}/f, 1/2)$ and any $\delta >0 $ such that $\delta \leq \alpha - c^{(q)}_{\max}/f$ and $\log (1/\delta) = \mathcal{O}(d)$. There exist universal constants $c_1, c_2$ such that if 
\begin{align*}
    \frac{n}{f} \geq c_1 C_\alpha^2 d \log (n/f),
\end{align*}
then with probability at least $1 - \exp (-f D(\alpha - \frac{c^{(q)}_{\max}}{f} ~||~ \delta))$, for all $t \geq 1$, the iterates of our algorithm with $\eta = \beta/(2\tilde{M}^2)$ satisfy

\begin{align}
    \norm{\bfw_t - \bfw^*} \leq  \left( \frac{1}{2} + \frac{1}{2} \sqrt{1 - \frac{\beta^2}{4\tilde{M}^2}}\right)^t \norm{\bfw_0 - \bfw^*} + c_2  \sqrt{\frac{d f}{n}}.  
\end{align}
\end{theorem}
An instance of a problem that satisfies the assumptions presented above is the linear regression problem. Formally, the data set consists of $n$ vectors $\{\mathbf{x}_1,\mathbf{x}_2,\dots,\mathbf{x}_n\}$, where $\forall i, \mathbf{x}_i\in\mathbb{R}^d$. We construct the data matrix $X$ of size $n\times d$ using these vectors as its rows.
The $n$ labels corresponding to the data points are computed as follows: $\mathbf{y} = X\mathbf{w}$, where $\mathbf{w}$ denotes the parameter set. For this problem, our loss function is the \emph{least-squares} loss, i.e., we have $l_i(\mathbf{w}) = \frac{1}{2} (\mathbf{y}_i - \mathbf{x}_i^T \mathbf{w})^2$ for $i=1, \dots, n$ where $\mathbf{x}_i^T$ denotes the $i^{\mathrm{th}}$ row of $X$. 


\subsection{Numerical Experiments}
\label{sec:linear_regression_dist_training}
We use the GD algorithm \eqref{eq:gd_update} with the initial randomly chosen parameter vector $\mathbf{w_0} \sim \mathcal{N}(\mathbf{0}_d, I_d)$. We partition the data matrix $X$ row-wise into $f$ submatrices $X_1,X_2,\dots,X_f$, and correspondingly the label vector $\mathbf{y}$ into $f$ sub-vectors  $\textbf{y}_1,\textbf{y}_2,\dots,\textbf{y}_f$, where $f$ is the number of files of the distributed algorithm. A file $B_{t,i}$ consists of a pair $(X_i, \textbf{y}_i)$. For each of its assigned files $B_{t_i} = (X_i, \textbf{y}_i) \in \mathcal{N}^w(U_i)$, worker $U_i$ either computes the honest partial gradient or a distorted value and returns it to the PS. Using the formulation of Section \ref{sec:all_attacks}, the gradient in Eq. \eqref{eq:returned_gradient} for linear regression is the product $\textbf{g}_{t,i} = X_i^TX_i\textbf{w} - X_i^T\textbf{y}_i$.

\textbf{Metrics}: For each scheme and value of $q$ we run multiple Carlo simulations, and calculated the average least-square loss that each algorithm converges to across the Monte Carlo simulations. 
For each simulation we declare convergence if the final empirical loss is less than 0.1 
We record the fraction of experiments that converged and the rate of convergence. In computing the average loss, the experiments that did not converged are not taken into account (for more details, please see Supplement Section \ref{appendix:monte_carlo_loss}).

\subsubsection{Experiment Setup}
In our experiments, we set $n=50000$, $d=100$ while our cluster consists of $K = 15$ workers. All replication-based schemes use $r = 3$. For Aspis+, we considered a $2-(15,3,1)$ design \cite{DRSCDCA}.

The geometric median is available as a Python library \cite{pillutla_geo_median}. Initially, we tuned the learning rate for each scheme and each distortion method to decide the one to use for the Monte Carlo simulations; all learning rates $10^{-x}$, $x=\{1,2,\dots,6\}$ have been tested. Also, we fix the random seeds of our experiments to be the same across all schemes; this guarantees that the data matrix as well as the original model estimate $\mathbf{w_0}$ will be the same across all methods. At the beginning of the algorithm, all elements of $X$ and $\mathbf{w}$ are generated randomly according to a $\mathcal{N}(0,1)$ distribution. For all runs, we chose to terminate the algorithm when the norm of gradient is less than $10^{-10}$ or the algorithm has reached a maximum number of 2000 iterations. Our code is available online \footnote{\href{https://github.com/kkonstantinidis/Aspis}{https://github.com/kkonstantinidis/Aspis}}.

\subsubsection{Results} 
\begin{figure}[t]
	\centering
	\begin{subfigure}[b]{0.43\textwidth}
		\centering
		\includegraphics[scale=0.4]{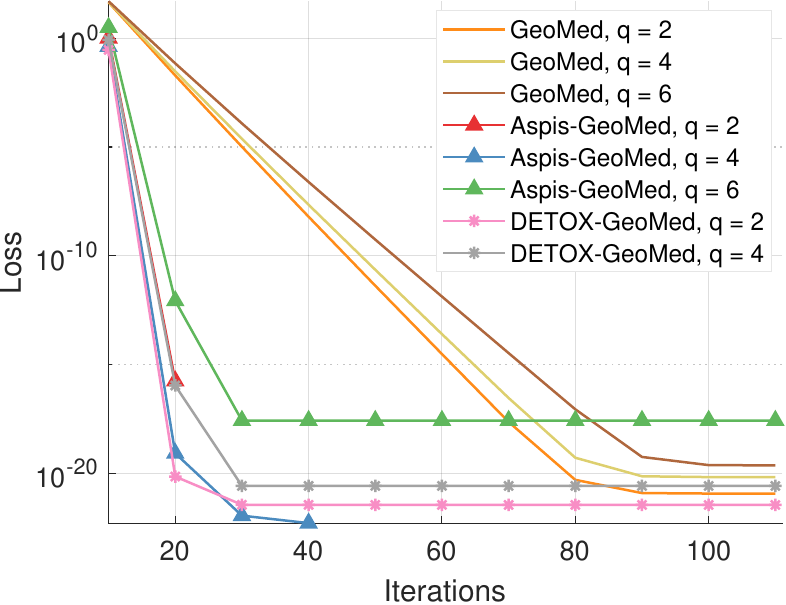}
		\caption{\emph{Reversed gradient} distortion.}
		\label{fig:loss_fig_127}
	\end{subfigure}
	\hspace{0.01\textwidth}
	\begin{subfigure}[b]{0.43\textwidth}
		\centering
		\includegraphics[scale=0.4]{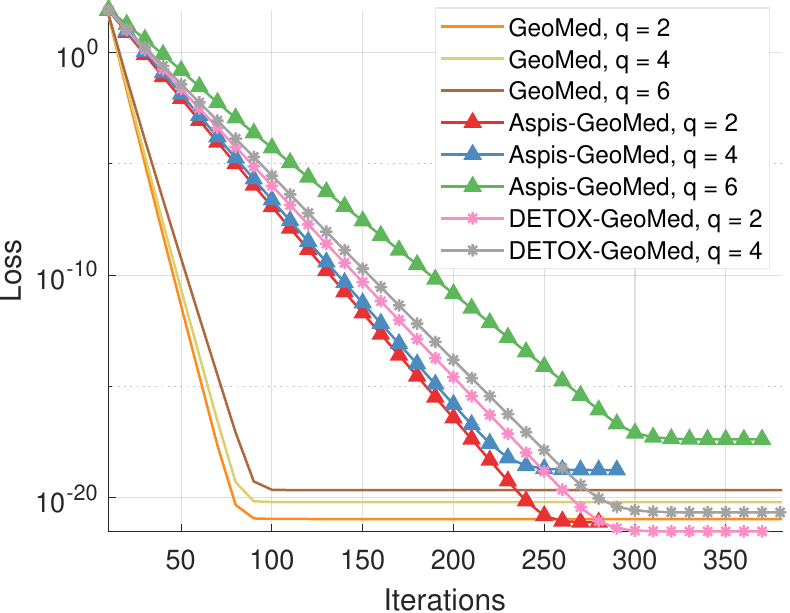}
		\caption{\emph{Constant} distortion.}
		\label{fig:loss_fig_128}
	\end{subfigure}
	\caption{Linear regression least-squares loss, optimal attacks, ATT-2 (Aspis), geometric median defenses, $K = 15$.}
\end{figure}

The first set of experiments are for the strong attack ATT-2. The baseline scheme where geometric median is applied on all $K$ gradients returned by the workers is referred to as \emph{GeoMed} and it has no redundancy. DETOX, Aspis, and Aspis+ use geometric median as part of the robust aggregation. Under reversed gradient (see Figure \ref{fig:loss_fig_127}), it is clear that all schemes perform well and achieve similar loss for $q = 2,4$ Byzantines. Nevertheless, baseline geometric median needed at least 100 iterations to converge while the redundancy-based schemes have a faster convergence rate. 
However, the situation is very different for $q = 6$ where Aspis converges within 30 iterations. In contrast, the DETOX loss diverged in all 100 simulations, while the convergence rate of the baseline scheme is much slower.

For the constant attack (see Fig. \ref{fig:loss_fig_128}) however, the relative performance of the baseline scheme and Aspis is reversed, i.e., the baseline scheme has a faster convergence rate as compared with Aspis. Moreover, Aspis and DETOX have roughly similar convergence rates for $q=2,4$. As before for $q=6$, none of the DETOX simulations converged. 

\begin{figure}
	\begin{center}
		\includegraphics[width=0.4\textwidth]{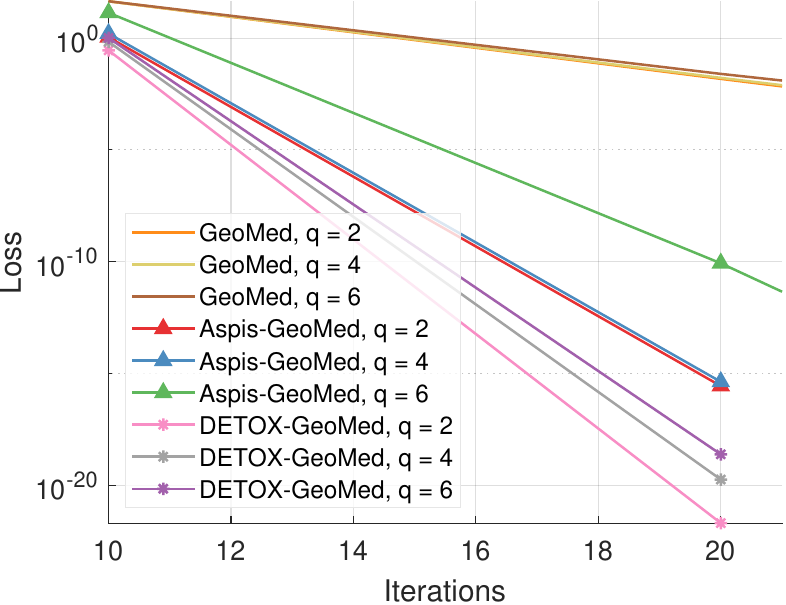}
	\end{center}
	\caption{Linear regression least-squares loss, \emph{ALIE} distortion, optimal attacks, ATT-2 (Aspis), geometric median defenses, $K = 15$.}
	\label{fig:loss_fig_126}
\end{figure}

Our last set of results are with the ALIE attack and the results are reported in Figure \ref{fig:loss_fig_126}. As the baseline geometric median simulations converged much slower than other schemes, they could not fit properly into the figure; it achieved final loss approximately equal to $2.07\times10^{-28}$, $9.71\times10^{-28}$, and $1.77\times10^{-28}$ for $q=2,4$, and $6$ respectively. On the other hand, Aspis converged to $1.69\times10^{-23}$ in 30 iterations for $q=6$. For this attack all schemes converged to very low loss values in all simulation runs. Nevertheless as is evident from Figure \ref{fig:loss_fig_126}, both Aspis and DETOX converge to loss values less that $10^{-5}$ within 15 iterations. 



\begin{figure}[t]
	\centering
	\begin{subfigure}[b]{0.43\textwidth}
		\centering
		\includegraphics[scale=0.4]{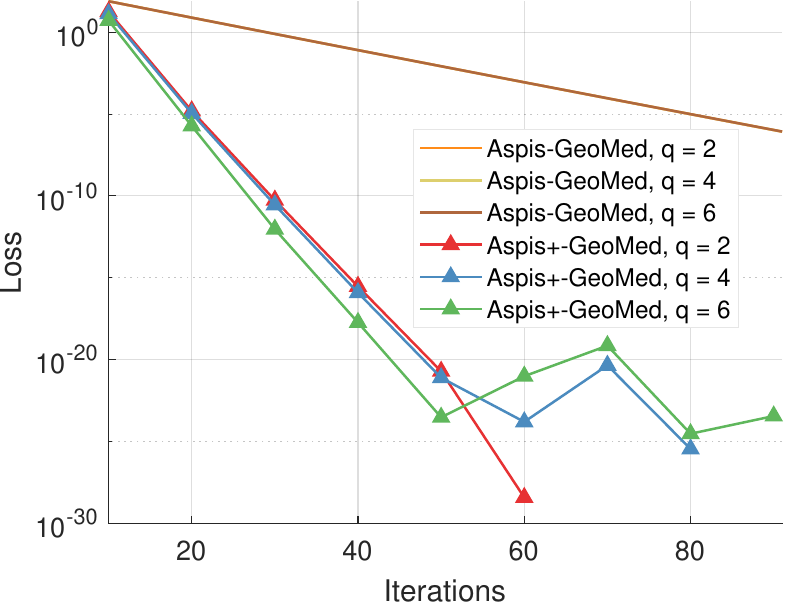}
		\caption{\emph{Reversed gradient} distortion.}
		\label{fig:loss_fig_129}
	\end{subfigure}
	\hspace{0.01\textwidth}
	\begin{subfigure}[b]{0.43\textwidth}
		\centering
		\includegraphics[scale=0.4]{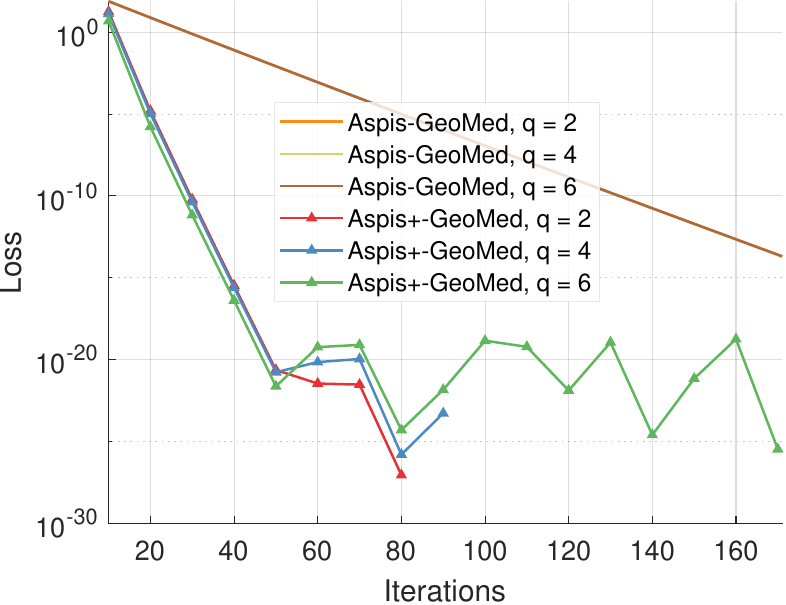}
		\caption{\emph{Constant} distortion.}
		\label{fig:loss_fig_130}
	\end{subfigure}
	\caption{Linear regression least-squares loss, random Byzantines (ATT-3), geometric median defenses, $K = 15$.}
\end{figure}

Another experiment we performed compares our two proposed methods, Aspis and Aspis+. 
For both schemes, we generate a new random Byzantine set $A$ every $T_b = 50$ iterations (introduced as ATT-3 Section \ref{sec:attack_3_aspis_plus}) while the detection window for Aspis+ is of length $T_d = 15$. For a comparable attack we use ATT-1 on Aspis (\emph{cf.} Section \ref{sec:attack_1_aspis}), i.e., all adversaries distort all their assigned files. We compare the two schemes under reversed gradient attack in Figure \ref{fig:loss_fig_129} and under constant attack in Figure \ref{fig:loss_fig_130}. Both methods achieve low final loss in the order of $10^{-20}$ or lower; Aspis converged to lower losses of the order of $10^{-24}$ in approximately 280 iterations in all cases. Nevertheless, Aspis+ achieves a faster convergence rate which aligns with the fact that it's mostly suitable for weaker adversaries.

\section{Distortion Fraction Evaluation}
\label{sec:distortion_fraction_analysis}
The main motivation of our distortion fraction analysis is that our deep learning experiments (\emph{cf.} Section \ref{sec:experiments}) and prior work \cite{byzshield} show that $\epsilon=c^{(q)}/f$ is a surrogate of the model's convergence with respect to accuracy. 
This comparison involves our work and state-of-the-art schemes under the best- and worst-case choice of the $q$ adversaries in terms of the achievable value of $\epsilon$. We also compare our work with \emph{baseline} approaches that do not involve redundancy or majority voting and aggregation is applied directly to the $K$ gradients returned by the workers ($f=K$, $c_{\mathrm{max}}^{(q)}=q$ and $\epsilon=q/K$).

For Aspis, we used the proposed attack ATT-2 from Section \ref{sec:attack_2_aspis} and the corresponding computation of $c^{(q), Aspis}$ of Theorem \ref{theorem:aspis_optimal_attack}. \emph{DETOX} in \cite{detox} employs a redundant assignment followed by majority voting and offers robustness guarantees which crucially rely on a ``random choice'' of the Byzantines. Our prior work \cite{byzshield} (ByzShield) has demonstrated the importance of a careful task assignment and observed that redundancy by itself is not sufficient to allow for Byzantine resilience. That work proposed an optimal choice of the $q$ Byzantines that maximizes $\epsilon^{DETOX}$, which we used in our current experiments. In short, DETOX splits the $K$ workers into $K/r$ groups. All workers within a group process the same subset of the batch, specifically containing $br/K$ samples. This phase is followed by majority voting on a group-by-group basis. Reference \cite{byzshield} suggests choosing the Byzantines so that at least $r'$ workers in each group are adversarial in order to distort the corresponding gradients. In this case, $c^{(q), DETOX} = \lfloor \frac{q}{r'} \rfloor$ and $\epsilon^{DETOX} = \lfloor \frac{q}{r'} \rfloor \times r/K$. We also compare with the distortion fraction incurred by ByzShield \cite{byzshield} under a worst-case scenario. For this scheme, there is no known optimal attack, and we performed an exhaustive combinatorial search to find the $q$ adversaries that maximize $\epsilon^{ByzShield}$ among all possible options; we follow the same process here to simulate ByzShield's distortion fraction computation while utilizing the scheme of that work based on \emph{mutually orthogonal Latin squares}. The reader can refer to Figure \ref{fig:epsilon_simulations_optimal_K50} and Appendix Tables \ref{table:epsilon_simulations_K15}, \ref{table:epsilon_simulations_K21}, and \ref{table:epsilon_simulations_K24} for our results. Aspis achieves major reductions in $\epsilon$; for instance, $\epsilon^{Aspis, ATT-2}$ is reduced by up to 99\% compared to both $\epsilon^{Baseline}$ and $\epsilon^{DETOX}$ in Figure \ref{fig:epsilon_simulations_optimal_K50}.

Next, we consider the \emph{weak} attack, ATT-1. For our scheme, we will make an arbitrary choice of $q$ adversaries which carry out the method introduced in Section \ref{sec:weak_subset_attack}, i.e., they will distort all files, and a successful detection is possible. As discussed in Section \ref{sec:weak_subset_attack}, the fraction of corrupted gradients is $\epsilon^{Aspis, ATT-1} = \binom{q}{r}/\binom{K}{r}$. For DETOX, a simple benign attack is used. To that end, let the $K/r$ files be $B_{t,0},B_{t,1},\dots,B_{t,K/r-1}$. Initialize $A=\emptyset$ and choose the $q$ Byzantines as follows: for $i=0,1,\dots,q-1$, among the remaining workers in $\{U_1,U_2,\dots,U_K\} - A$ add a worker from the group $B_{t,i \mod K/r}$ to the adversarial set $A$. Then,
\begin{equation*}
	c^{(q), DETOX} = \left\{
	\begin{array}{ll}
		q-\frac{K}{r}(r'-1) & \text{if } q > \frac{K}{r}(r'-1),\\
		0 & \text{otherwise}.
	\end{array}
	\right.
\end{equation*}
The results of this scenario are in Figure \ref{fig:epsilon_simulations_weak_K50}. 



\begin{figure*}[t]
	\centering
	\begin{subfigure}[b]{0.32\textwidth}
		\centering
		\includegraphics[scale=0.4]{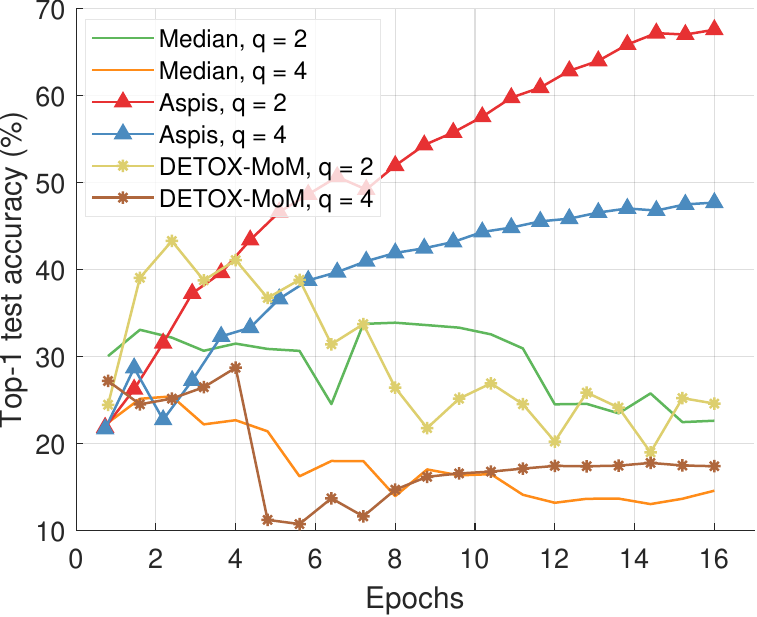}
		\caption{Median-based defenses.}
		\label{fig:top1_fig_94}
	\end{subfigure}
	\hfill
	\begin{subfigure}[b]{0.32\textwidth}
		\centering
		\includegraphics[scale=0.4]{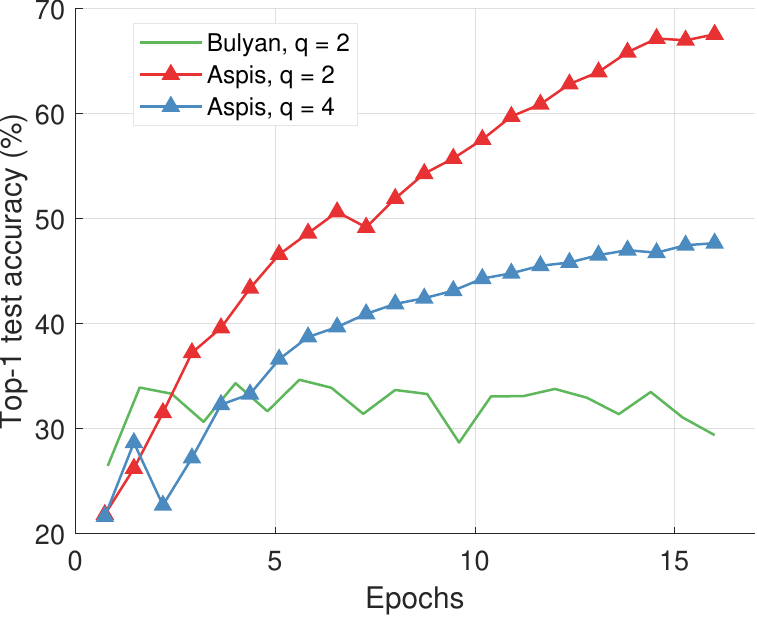}
		\caption{\emph{Bulyan}-based defenses.}
		\label{fig:top1_fig_95}
	\end{subfigure}
	\hfill
	\begin{subfigure}[b]{0.32\textwidth}
		\centering
		\includegraphics[scale=0.4]{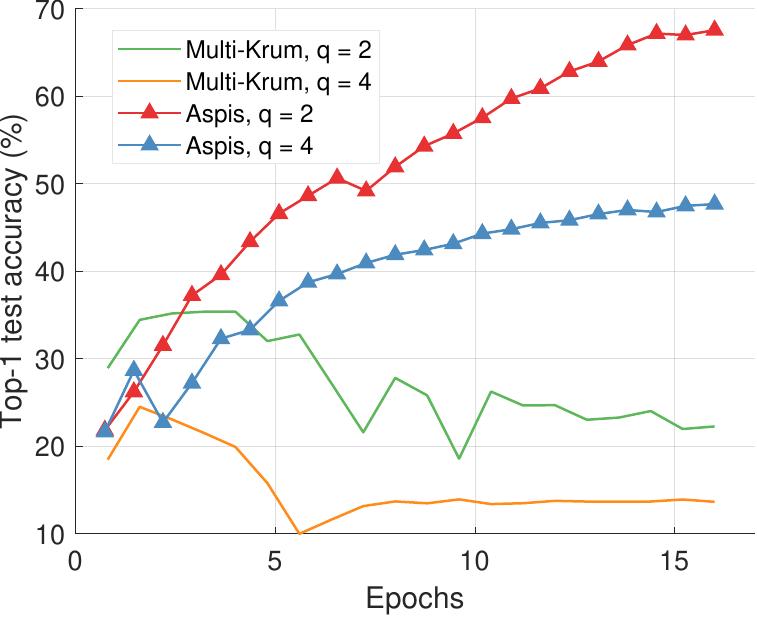}
		\caption{\emph{Multi-Krum}-based defenses.}
		\label{fig:top1_fig_96}
	\end{subfigure}
	\caption{\emph{ALIE} distortion under optimal attack scenarios, ATT-2 for Aspis, CIFAR-10, $K=15$.}
	\label{fig:top1_ALIE_optimal}
\end{figure*}

\begin{figure*}[t]
	\centering
	\begin{subfigure}[b]{0.32\textwidth}
		\centering
		\includegraphics[scale=0.4]{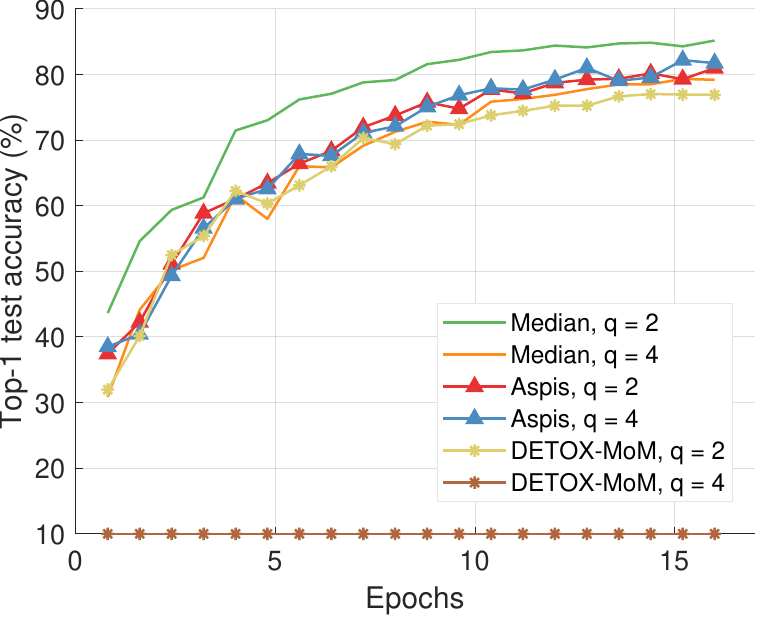}
		\caption{Median-based defenses.}
		\label{fig:top1_fig_97}
	\end{subfigure}
	\hfill
	\begin{subfigure}[b]{0.32\textwidth}
		\centering
		\includegraphics[scale=0.4]{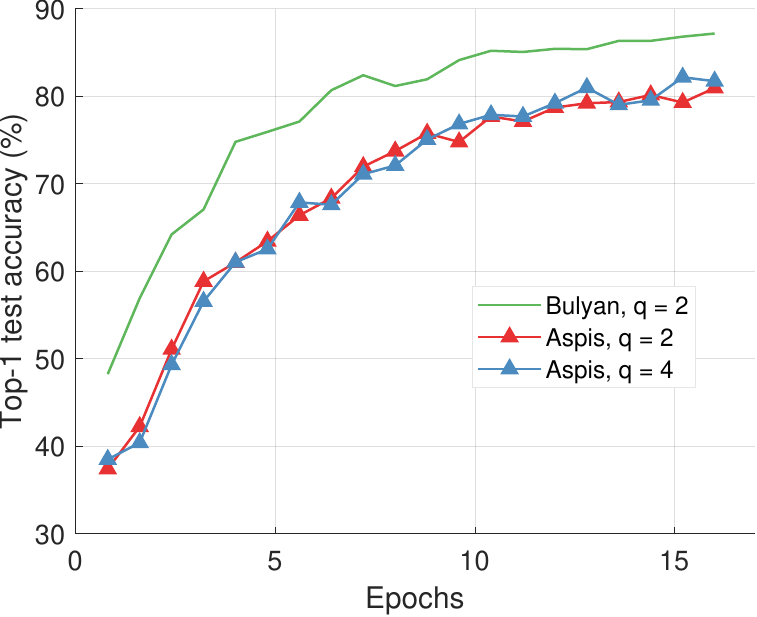}
		\caption{\emph{Bulyan}-based defenses.}
		\label{fig:top1_fig_98}
	\end{subfigure}
	\hfill
	\begin{subfigure}[b]{0.32\textwidth}
		\centering
		\includegraphics[scale=0.4]{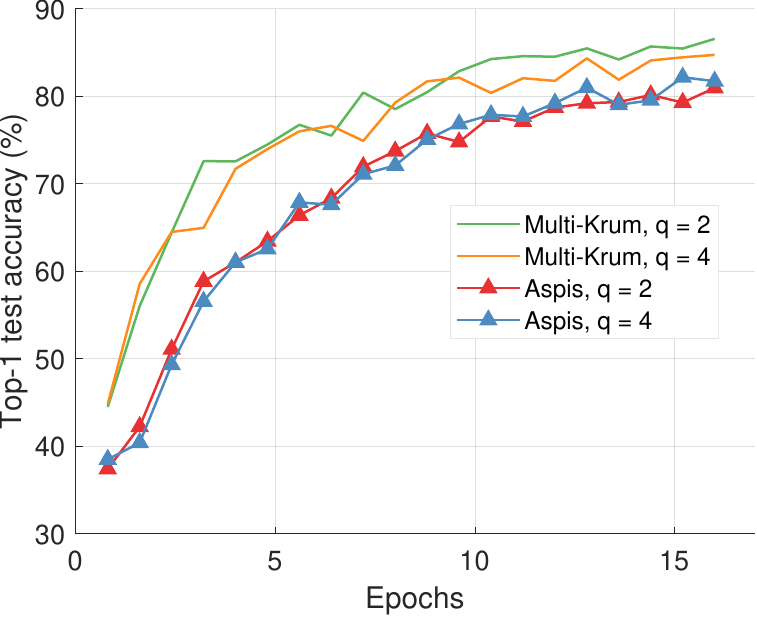}
		\caption{\emph{Multi-Krum}-based defenses.}
		\label{fig:top1_fig_99}
	\end{subfigure}
	\caption{\emph{Reversed gradient} distortion under optimal attack scenarios, ATT-2 for Aspis, CIFAR-10, $K=15$.}
\end{figure*}

\section{Large-Scale Deep Learning Experiments}
\label{sec:deep_learning_experiments}
All these experiments are performed under Setting-I, i.e., no assumptions are made about the dataset or the loss function. Accordingly, the evaluation here is in terms of the distortion fraction (see Section \ref{sec:distortion_fraction_analysis}) and numerical experiments (described below). For the experiments, we used the mini-batch SGD (see \eqref{eq:vanilla_sgd_update}) and the robust estimator (see Algorithm \ref{alg:main_algorithm}) is the coordinate-wise median. 

\subsection{Experiment Setup}
\label{sec:experiment_setup}
We have evaluated the performance of our methods and competing techniques in classification tasks on Amazon EC2 clusters. The project is written in PyTorch \cite{pytorch} and uses the MPICH library for communication between the different nodes. We worked with the CIFAR-10 data set \cite{cifar10} using the ResNet-18 \cite{he_resnet} model. We used clusters of $K = 15$, $21$, $25$ workers, redundancy $r = 3$, and simulated values of $q=2,4,6,7,9$ during training.
Detailed information about the implementation can be found in Appendix Section \ref{appendix:implementation_details}.


{\bf Competing methods:} We compare Aspis against the baseline implementations of median-of-means \cite{minsker2015}, Bulyan \cite{bulyan}, and Multi-Krum \cite{blanchard_krum}. If $c_{\mathrm{max}}^{(q)}$ is the number of adversarial computations, 
then Bulyan requires at least $4c_{\mathrm{max}}^{(q)}+3$ total number of computations while the same number for Multi-Krum is $2c_{\mathrm{max}}^{(q)}+3$. These constraints make these methods inapplicable for larger values of $q$ for which our methods are robust. The second class of comparisons is with methods that use redundancy, specifically DETOX \cite{detox}. For the baseline scheme we compare with median-based techniques since they originate from robust statistics and are the basis for many aggregators. Multi-Krum combines the intuitions of majority-based and squared-distance-based methods. Draco \cite{draco} is a closely related method that uses redundancy. However we do not compare with it since it is very limited in the number of Byzantines that it is resilient to. 


Note that for a baseline scheme, all choices of $A$ are equivalent in terms of the value of $\epsilon$. In our comparisons between Aspis and DETOX we will consider two attack scenarios concerning the choice of the adversaries. For the \emph{optimal} attack on DETOX, we will use the method proposed in \cite{byzshield} and compare with the attack introduced in Section \ref{sec:optimal_subset_attack}. For the \emph{weak} one, we will choose the adversaries such that they incur the minimum value of $\epsilon$ in DETOX for given $q$ and compare its performance with the scenario of Section \ref{sec:weak_subset_attack}. All schemes compared with Aspis+ consider random sets of Byzantines, and for Aspis+, we will use the attack ATT-3.

\begin{figure*}[!htb]
	\minipage{0.32\textwidth}
	\includegraphics[scale=0.4]{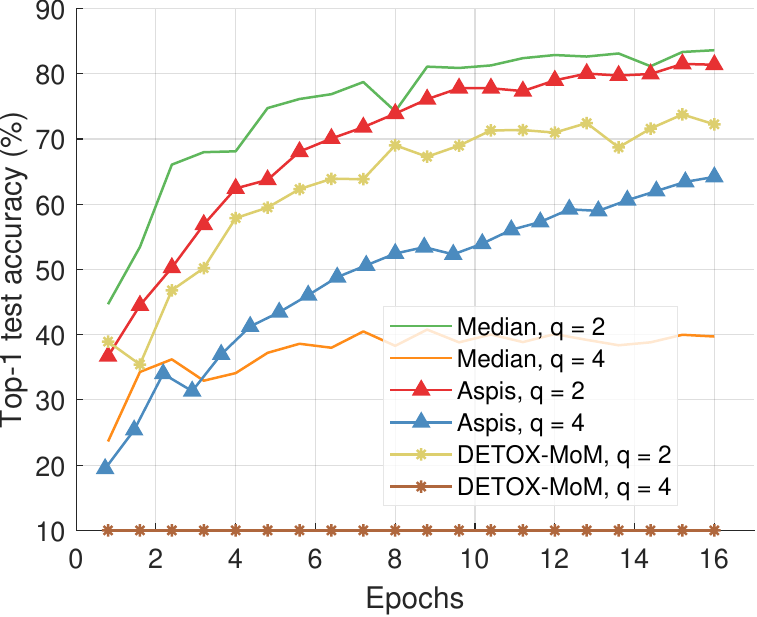}
	\caption{\emph{FoE} distortion, optimal attacks, ATT-2 (Aspis) and median-based defenses (CIFAR-10), $K=15$}
	\label{fig:top1_fig_101}
	\endminipage\hfill
	\minipage{0.32\textwidth}
	\includegraphics[scale=0.4]{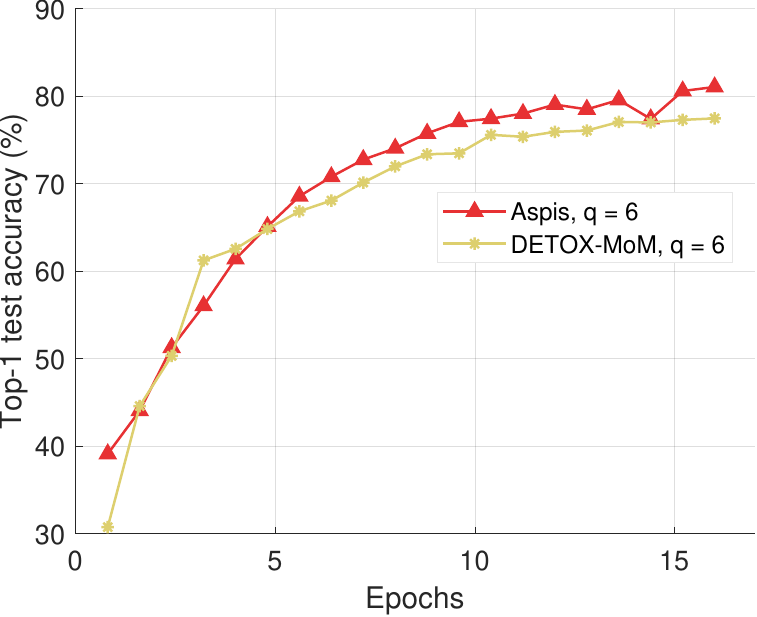}
	\caption{\emph{Reversed gradient} distortion, weak attacks, ATT-1 (Aspis) and median-based defenses (CIFAR-10), $K=15$.}
	\label{fig:top1_fig_100}
	\endminipage\hfill
	\minipage{0.32\textwidth}
	\includegraphics[scale=0.4]{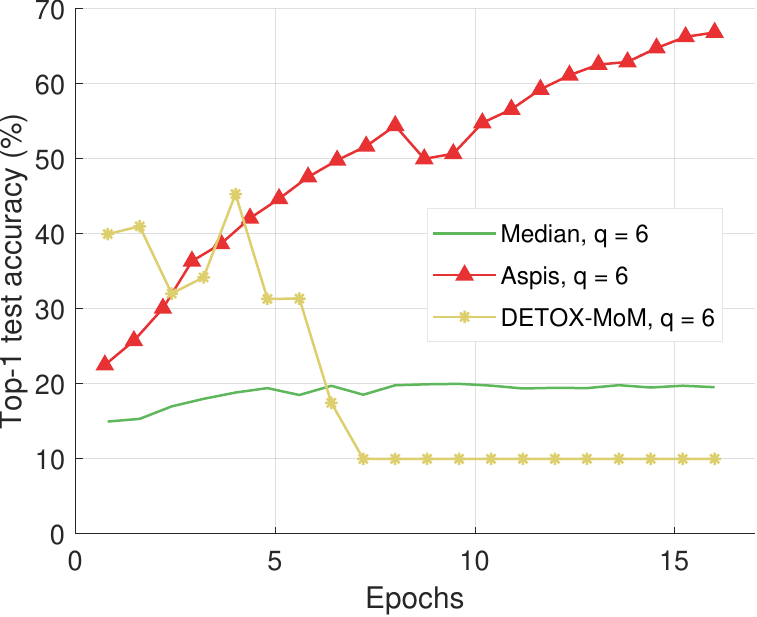}
	\caption{\emph{ALIE} distortion, weak attacks, ATT-1 (Aspis) and median-based defenses (CIFAR-10), $K=15$.}
	\label{fig:top1_fig_88}
	\endminipage
\end{figure*}

\begin{figure*}[t]
	\centering
	\begin{subfigure}[b]{0.32\textwidth}
		\centering
		\includegraphics[scale=0.4]{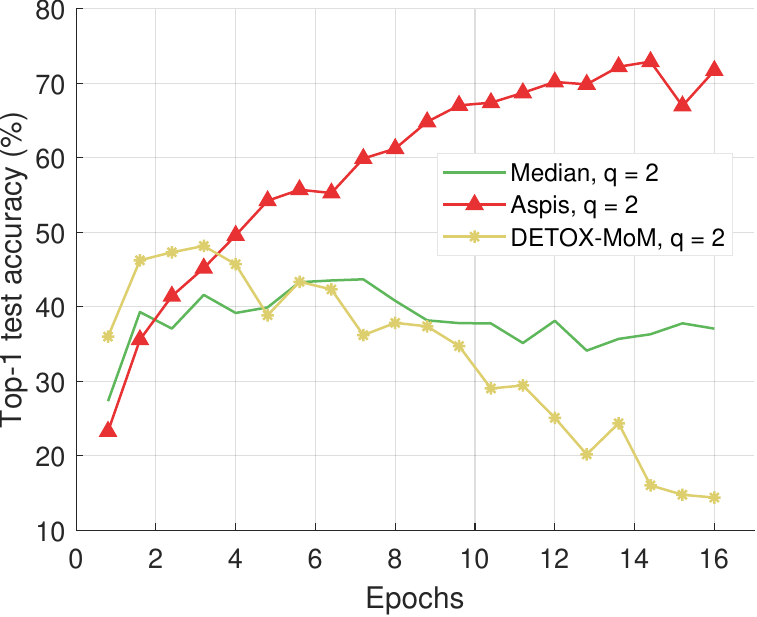}
		\caption{Median-based defenses.}
		\label{fig:top1_fig_102}
	\end{subfigure}
	\hfill
	\begin{subfigure}[b]{0.32\textwidth}
		\centering
		\includegraphics[scale=0.4]{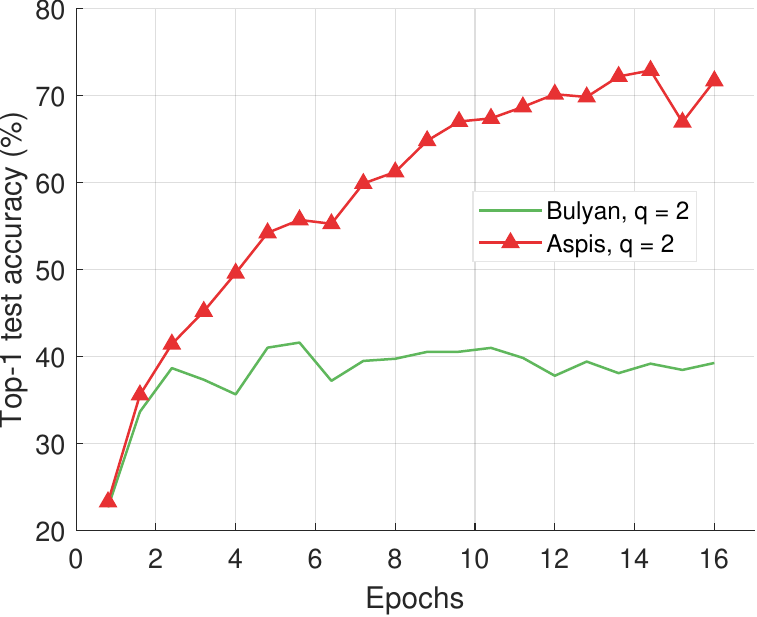}
		\caption{\emph{Bulyan}-based defenses.}
		\label{fig:top1_fig_103}
	\end{subfigure}
	\hfill
	\begin{subfigure}[b]{0.32\textwidth}
		\centering
		\includegraphics[scale=0.4]{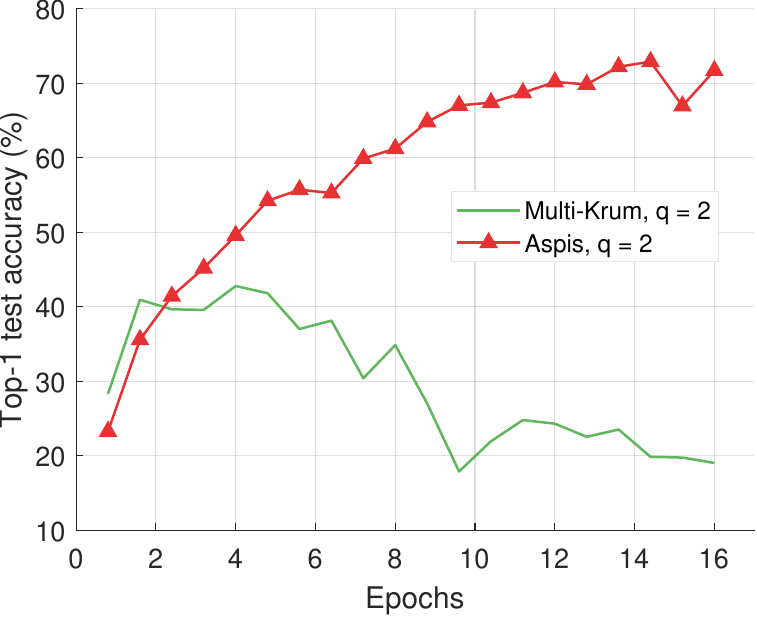}
		\caption{\emph{Multi-Krum}-based defenses.}
		\label{fig:top1_fig_104}
	\end{subfigure}
	\caption{\emph{ALIE} distortion under optimal attack scenarios, ATT-2 for Aspis, CIFAR-10, $K=21$.}
	\label{fig:top1_ALIE_optimal_K21}
\end{figure*}

\subsection{Aspis Experimental Results}
\label{sec:experiments}


\subsubsection{Comparison under Optimal Attacks}
We compare the different defense algorithms under optimal attack scenarios using ATT-2 for Aspis. Figure \ref{fig:top1_fig_94} compares our scheme Aspis with the baseline implementation of coordinate-wise median ($\epsilon=0.133, 0.267$ for $q=2,4$, respectively) and DETOX with median-of-means ($\epsilon=0.2, 0.4$ for $q=2,4$, respectively) under the ALIE attack. Aspis converges faster and achieves at least a 35\% average accuracy boost (at the end of the training) for both values of $q$ ($\epsilon^{Aspis}=0.004, 0.062$ for $q=2,4$, respectively).\footnote{Please refer to Appendix Tables \ref{table:epsilon_simulations_K15} and \ref{table:epsilon_simulations_K21} for the values of the distortion fraction $\epsilon$ each scheme incurs.} In Figures \ref{fig:top1_fig_95} and \ref{fig:top1_fig_96}, we observe similar trends in our experiments with Bulyan and Multi-Krum, where Aspis significantly outperforms these techniques. For the current setup, Bulyan is not applicable for $q=4$ since $K = 15 < 4c_{\mathrm{max}}^{(q)}+3 = 4q+3 = 19$. Also, neither Bulyan nor Multi-Krum can be paired with DETOX for $q \geq 4$ since the inequalities $f \geq 4c_{\mathrm{max}}^{(q)}+3$ and $f \geq 2c_{\mathrm{max}}^{(q)}+3$, where $f = f_{\mathrm{DETOX}} = K/r$, cannot be satisfied; for the specific case of Bulyan even $q=2,3$ would not be supported by DETOX. Please refer to Section \ref{sec:experiment_setup} and Section \ref{sec:distortion_fraction_analysis} for more details on these requirements. Also, note that the accuracy of most competing methods fluctuates more than in the results presented in the corresponding papers \cite{detox} and \cite{alie}. This is expected as we consider stronger attacks than those papers, i.e., optimal deterministic attacks on DETOX and, in general, up to 27\% adversarial workers in the cluster. Also, we have done multiple experiments with different random seeds to demonstrate the stability and superiority of our accuracy results compared to other methods (against median-based defenses in Appendix Figure \ref{fig:top1_error_bar_median}, Bulyan in Figure \ref{fig:top1_error_bar_bulyan} and Multi-Krum in Supplement Figure \ref{fig:top1_error_bar_multikrum}); we point the reader to Appendix Section \ref{appendix:error_bars} for this analysis. This analysis is clearly missing from most prior work, including that of ALIE \cite{alie} and their presented results are only a snapshot of a single experiment. The results for the reversed gradient attack are shown in Figures \ref{fig:top1_fig_97}, \ref{fig:top1_fig_98}, and \ref{fig:top1_fig_99}. Given that this is a weaker attack \cite{byzshield,detox} all schemes, including the baseline methods, are expected to perform well; indeed, in most cases, the model converges to approximately 80\% accuracy. However, DETOX fails to converge to high accuracy for $q=4$ as in the case of ALIE; one explanation is that $\epsilon^{DETOX}=0.4$ for $q=4$. Under the Fall of Empires (FoE) distortion (\emph{cf.} Figure \ref{fig:top1_fig_101}) our method still enjoys an accuracy advantage over the baseline and DETOX schemes which becomes more important as the number of Byzantines in the cluster increases.


We have also performed experiments on larger clusters ($K=21$ workers) as well. The results for the ALIE distortion with the ATT-2 attack can be found in Figure \ref{fig:top1_ALIE_optimal_K21}. They exhibit similar behavior as in the case of $K=15$.

\begin{figure}
	\begin{center}
		\includegraphics[width=0.4\textwidth]{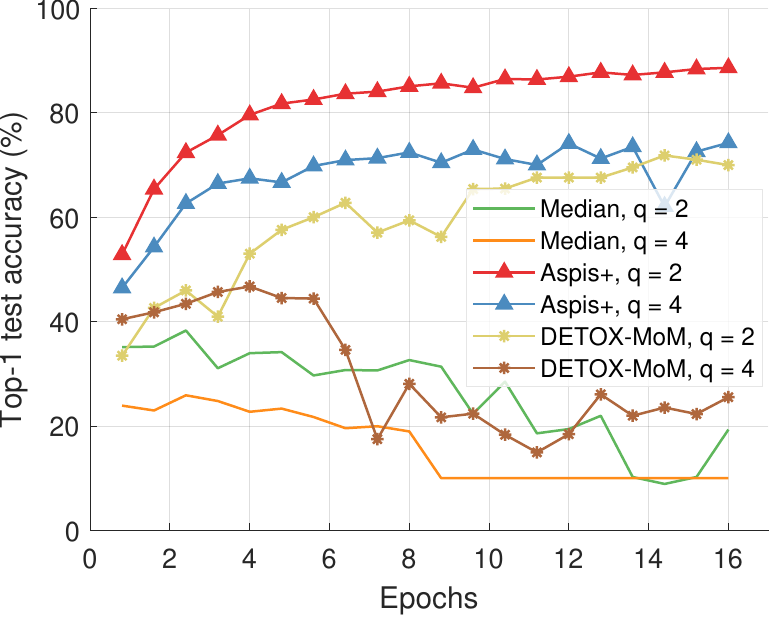}
	\end{center}
	\caption{\emph{ALIE} distortion and random Byzantines, $K=15$ (median-based defenses). ATT-3 used on Aspis+.}
	\label{fig:top1_fig_111}
\end{figure}

\begin{figure}
	\begin{center}
		\includegraphics[width=0.4\textwidth]{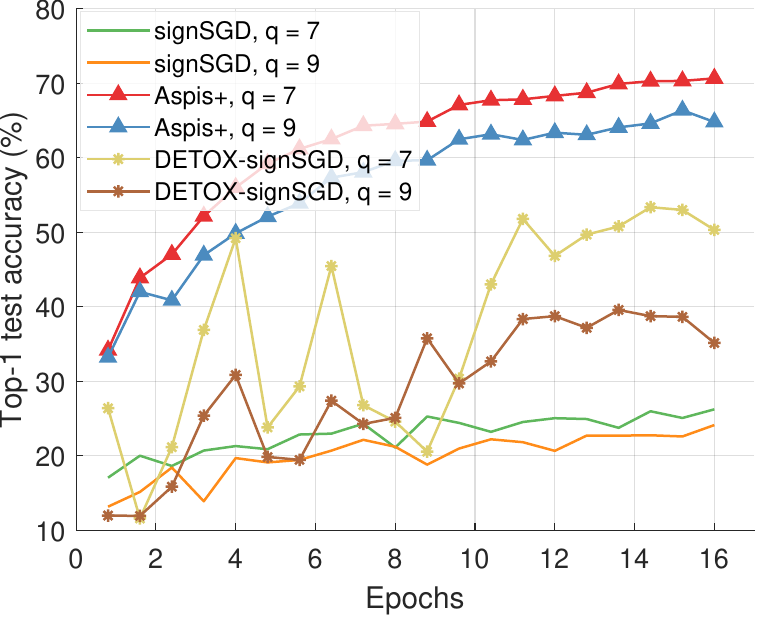}
	\end{center}
	\caption{\emph{Constant} distortion and random Byzantines, $K=25$ (\emph{signSGD}-based defenses). ATT-3 used on Aspis+.}
	\label{fig:top1_fig_112}
\end{figure}

\subsubsection{Comparison under Weak Attacks}
For baseline schemes, the discussion of weak versus optimal choice of the adversaries is not very relevant as any choice of the $q$ Byzantines can overall distort exactly $q$ out of the $K$ gradients. Hence, for weak scenarios, we chose to compare mostly with DETOX while using ATT-1 on Aspis. The accuracy is reported in Figures \ref{fig:top1_fig_100} and \ref{fig:top1_fig_88}, according to which Aspis shows an improvement under attacks on the more challenging end of the spectrum (ALIE). According to Appendix Table \ref{table:epsilon_simulations_weak_K15}, Aspis enjoys a fraction $\epsilon^{Aspis} = 0.044$ while $\epsilon^{Baseline} = 0.4$ and $\epsilon^{DETOX} = 0.2$ for $q=6$.


\subsection{Aspis+ Experimental Results}
\label{sec:experiments_aspis+}
For Aspis+, we considered the attack ATT-3 discussed in Section \ref{sec:attack_3_aspis_plus}. We tested clusters of $K=15$ with $q=2,4$ and $K=25$ workers among which $q=7,9$ are Byzantine. In the former case, a $2-(15,3,1)$ design \cite{DRSCDCA} with $f=35$ blocks (files) was used for the placement, while in the latter case, we used a $2-(25,3,1)$ design \cite{DRSCDCA} with $f=100$ blocks (files). A new random Byzantine set $A$ is generated every $T_b = 50$ iterations while the detection window is of length $T_d = 15$.

The results for $K=15$ are in Figure \ref{fig:top1_fig_111}. We tested against the ALIE distortion, and all compared methods use median-based defenses to filter the gradients. Aspis+ demonstrates an advantage of at least 15\% compared with other algorithms (\emph{cf.} $q=2$). For $K=25$, we tried a weaker distortion than ALIE, i.e., the constant attack paired with \emph{signSGD}-based defenses \cite{SIGNSGD}. In signSGD, the PS will output the majority of the gradients' signs for each dimension. Following the advice of \cite{detox}, we pair this defense with the stronger constant attack as sign flips (\emph{e.g.,} reversed gradient) are unlikely to affect the gradient's distribution. Aspis+ with median still enjoys an accuracy improvement of at least 20\% for $q=7$ and a larger one for $q=9$. The results are in Figure \ref{fig:top1_fig_112}; in this figure, the DETOX accuracy is an average of two experiments using two different random seeds.

\section{Conclusions and Future Work}
In this work, we have presented Aspis and Aspis+, two Byzantine-resilient distributed schemes that use redundancy and robust aggregation in novel ways to detect failures of the workers. Our theoretical analysis and numerical experiments clearly indicate their superior performance compared to state-of-the-art. Our experiments show that these methods require increased computation and communication time as compared to prior work, e.g., note that each worker has to transmit $l$ gradients instead of $1$ in related work \cite{detox,draco} (see Appendix Section \ref{appendix:computation_communication_overhead} for details). We emphasize, however, that our schemes converge to high accuracy in our experiments, while other methods remain at much lower accuracy values regardless of how long the algorithm runs for.


Our experiments involve clusters of up to $25$ workers. As we scale Aspis to more workers, the total number of files and the computation load $l$ of each worker will also scale; this increases the memory needed to store the gradients during aggregation. For complex neural networks, the memory to store the model and the intermediate gradient computations is by far the most memory-consuming aspect of the algorithm. For these reasons, Aspis is mostly suitable for training large data sets using fairly compact models that do not require too much memory. Aspis+, on the other hand, is a good fit for clusters that suffer from non-adversarial failures that can lead to inaccurate gradients. Finally, utilizing GPUs and communication-related algorithmic improvements are worth exploring to reduce the time overhead.

\section{Appendix}
\begin{table}[!ht]
	\centering
	\caption{Main notation of the paper.}
	\label{table:notation}
	\resizebox{1\columnwidth}{!}{
		\begin{tabular}{P{1cm}p{6cm}}
			\hline
			Symbol & Meaning \\
			\hline
			$K$ & number of workers\\
			$q$ & number of adversaries\\
			$r$ & redundancy (number of workers each file is assigned to)\\
			$b$ & batch size\\
			$B_t$ & samples of batch of $t^{\mathrm{th}}$ iteration\\
			$f$ & number of files (alternatively called \emph{groups} or \emph{tasks})\\
			$U_j$ & $j^{\mathrm{th}}$ worker\\
			$l$ & computation load (number of files per worker)\\
			$\calN^w(U_j)$ & set of files of worker $U_j$\\
			$\calN^f(B_{t,i})$ & set of workers assigned to file $B_{t,i}$\\
			$\mathbf{g}_{t,i}$ & true gradient of file $B_{t,i}$ with respect to $\mathbf{w}$\\
			$\hat{\mathbf{g}}_{t,i}^{(j)}$ & returned gradient of $U_j$ for file $B_{t,i}$ with respect to $\mathbf{w}$\\
			$\mathbf{m}_i$ & majority gradient for file $B_{t,i}$\\
			$\calU$ & worker set $\{U_1,U_2,...,U_K\}$\\
			$\mathbf{G}_{task}$ & graph used to encode the task assignments to workers\\
			$\mathbf{G}_t$ & graph indicating the agreements of pairs of workers in all of their common gradient tasks in $t^{\mathrm{th}}$ iteration\\
			$A$ & set of adversaries\\
			$M_{\mathbf{G}}$ & maximum clique in $\mathbf{G}$\\
			$c^{(q)}$ & number of distorted gradients after detection and aggregation\\
			$c_{\mathrm{max}}^{(q)}$ & maximum number of distorted gradients after detection and aggregation (worst-case)\\
			$D_i$ & disagreement set (of workers) for $i^{\mathrm{th}}$ adversary\\
			$r'$ & $(r+1)/2$, i.e., minimum number of distorted copies needed to corrupt majority vote for a file\\
			$\epsilon$ & $c^{(q)}/f$, i.e., fraction of distorted gradients after detection and aggregation\\
			$X_j$ & subset of files where the set of active adversaries is of size $j$; for linear regression this is the data matrix corresponding to the $i^\mathrm{th}$ file\\
			$X$ & data matrix of linear regression\\
			$n$ & number of points of linear regression\\
			$d$ & dimensionality of linear regression model\\
			\hline
		\end{tabular}
		}
\end{table}

\begin{table*}[!t]
	\large
	\newcommand\Ksubtablewidth{0.43\linewidth}
	\newcommand\Kresizetabular{\columnwidth}
	\centering
	\captionsetup[subtable]{position = below}
	\caption{Distortion fraction of optimal and weak attacks for $(K,f,l,r)=(15,455,91,3)$ and comparison.}
	\label{table:epsilon_simulations_K15}
	\begin{subtable}{\Ksubtablewidth}
		\centering
		{
			\resizebox{0.95\Kresizetabular}{!}{
				\begin{tabular}{P{0.5cm} R R R R}
					\toprule
					\multicolumn{1}{c}{$q$} & \multicolumn{1}{c}{$\epsilon^{Aspis}_{ATT-2}$} & \multicolumn{1}{c}{$\epsilon^{Baseline}$} & \multicolumn{1}{c}{$\epsilon^{DETOX}$} & \multicolumn{1}{c}{$\epsilon^{ByzShield}$} \\
					\hline
					$2$ & 0.004 & 0.133 & 0.2 & 0.04 \\
					$3$ & 0.022 & 0.2 & 0.2 & 0.12 \\
					$4$ & 0.062 & 0.267 & 0.4 & 0.2 \\
					$5$ & 0.132 & 0.333 & 0.4 & 0.32 \\
					$6$ & 0.242 & 0.4 & 0.6 & 0.48 \\
					$7$ & 0.4 & 0.467 & 0.6 & 0.56 \\
					\toprule
				\end{tabular}
			}
			\caption{Optimal attacks.}
		}
	\end{subtable}
	\hspace{0.01\textwidth}
	\begin{subtable}{\Ksubtablewidth}
		\centering
		{
			\resizebox{0.72\Kresizetabular}{!}{
				\begin{tabular}{P{0.5cm} R R R}
					\toprule
					$q$ & \multicolumn{1}{c}{$\epsilon^{Aspis}_{ATT-1}$} & \multicolumn{1}{c}{$\epsilon^{Baseline}$} & \multicolumn{1}{c}{$\epsilon^{DETOX}$} \\
					\hline
					$2$ & 0.002 & 0.133 & 0 \\
					$3$ & 0.002 & 0.2 & 0 \\
					$4$ & 0.009 & 0.267 & 0 \\
					$5$ & 0.022 & 0.333 & 0 \\
					$6$ & 0.044 & 0.4 & 0.2 \\
					$7$ & 0.077 & 0.467 & 0.4 \\
					\toprule
				\end{tabular}
			}
			\caption{Weak attacks.}
			\label{table:epsilon_simulations_weak_K15}
		}
	\end{subtable}
\end{table*}

\begin{table*}[!t]
	\large
	\newcommand\Ksubtablewidth{0.43\linewidth}
	\newcommand\Kresizetabular{\columnwidth}
	\centering
	\captionsetup[subtable]{position = below}
	\caption{Distortion fraction of optimal and weak attacks for $(K,f,l,r)=(21,1330,190,3)$ and comparison.}
	\label{table:epsilon_simulations_K21}
	\begin{subtable}{\Ksubtablewidth}
		\centering
		{
			\resizebox{0.95\Kresizetabular}{!}{
				\begin{tabular}{P{0.6cm} R R R R}
					\toprule
					$q$ & \multicolumn{1}{c}{$\epsilon^{Aspis}_{ATT-2}$} & \multicolumn{1}{c}{$\epsilon^{Baseline}$} & \multicolumn{1}{c}{$\epsilon^{DETOX}$} & \multicolumn{1}{c}{$\epsilon^{ByzShield}$} \\
					\hline
					$2$ & 0.002 & 0.095 & 0.143 & 0.02 \\
					$3$ & 0.008 & 0.143 & 0.143 & 0.06 \\
					$4$ & 0.021 & 0.19 & 0.286 & 0.1 \\
					$5$ & 0.045 & 0.238 & 0.286 & 0.16 \\
					$6$ & 0.083 & 0.286 & 0.429 & 0.24 \\
					$7$ & 0.137 & 0.333 & 0.429 & 0.33 \\
					$8$ & 0.211 & 0.381 & 0.571 & 0.43 \\
					$9$ & 0.307 & 0.429 & 0.571 & 0.51 \\
					$10$ & 0.429 & 0.476 & 0.714 & 0.59 \\
					\toprule
				\end{tabular}
			}
			\caption{Optimal attacks.}
		}
	\end{subtable}
	\hspace{0.01\textwidth}
	\begin{subtable}{\Ksubtablewidth}
		\centering
		{
			\resizebox{0.72\Kresizetabular}{!}{
				\begin{tabular}{P{0.6cm} R R R}
					\toprule
					$q$ & \multicolumn{1}{c}{$\epsilon^{Aspis}_{ATT-1}$} & \multicolumn{1}{c}{$\epsilon^{Baseline}$} & \multicolumn{1}{c}{$\epsilon^{DETOX}$} \\
					\hline
					$2$ & 0.001 & 0.095 & 0 \\
					$3$ & 0.001 & 0.143 & 0 \\
					$4$ & 0.003 & 0.19 & 0 \\
					$5$ & 0.008 & 0.238 & 0 \\
					$6$ & 0.015 & 0.286 & 0 \\
					$7$ & 0.026 & 0.333 & 0 \\
					$8$ & 0.042 & 0.381 & 0.143 \\
					$9$ & 0.063 & 0.429 & 0.286 \\
					$10$ & 0.09 & 0.476 & 0.429 \\
					\toprule
				\end{tabular}
			}
			\caption{Weak attacks.}
		}
	\end{subtable}
\end{table*}

\begin{table*}[!t]
	\large
	\newcommand\Ksubtablewidth{0.43\linewidth}
	\newcommand\Kresizetabular{\columnwidth}
	\centering
	\captionsetup[subtable]{position = below}
	\caption{Distortion fraction of optimal and weak attacks for $(K,f,l,r)=(24,2024,253,3)$ and comparison.}
	\label{table:epsilon_simulations_K24}
	\begin{subtable}{\Ksubtablewidth}
		\centering
		{
			\resizebox{0.95\Kresizetabular}{!}{
				\begin{tabular}{P{0.6cm} R R R R}
					\toprule
					$q$ & \multicolumn{1}{c}{$\epsilon^{Aspis}_{ATT-2}$} & \multicolumn{1}{c}{$\epsilon^{Baseline}$} & \multicolumn{1}{c}{$\epsilon^{DETOX}$} & \multicolumn{1}{c}{$\epsilon^{ByzShield}$} \\
					\hline
					$2$ & 0.001 & 0.083 & 0.125 & 0.031 \\
					$3$ & 0.005 & 0.125 & 0.125 & 0.063 \\
					$4$ & 0.014 & 0.167 & 0.25 & 0.125 \\
					$5$ & 0.03 & 0.208 & 0.25 & 0.188 \\
					$6$ & 0.054 & 0.25 & 0.375 & 0.281 \\
					$7$ & 0.09 & 0.292 & 0.375 & 0.375 \\
					$8$ & 0.138 & 0.333 & 0.5 & 0.5 \\
					$9$ & 0.202 & 0.375 & 0.5 & 0.5 \\
					$10$ & 0.282 & 0.417 & 0.625 & 0.531 \\
					$11$ & 0.38 & 0.458 & 0.625 & 0.625\\
					\toprule
				\end{tabular}
			}
			\caption{Optimal attacks.}
		}
	\end{subtable}
	\hspace{0.01\textwidth}
	\begin{subtable}{\Ksubtablewidth}
		\centering
		{
			\resizebox{0.72\Kresizetabular}{!}{
				\begin{tabular}{P{0.6cm} R R R}
					\toprule
					$q$ & \multicolumn{1}{c}{$\epsilon^{Aspis}_{ATT-1}$} & \multicolumn{1}{c}{$\epsilon^{Baseline}$} & \multicolumn{1}{c}{$\epsilon^{DETOX}$} \\
					\hline
					$2$ & 0 & 0.083 & 0 \\
					$3$ & 0 & 0.125 & 0 \\
					$4$ & 0.002 & 0.167 & 0 \\
					$5$ & 0.005 & 0.208 & 0 \\
					$6$ & 0.01 & 0.25 & 0 \\
					$7$ & 0.017 & 0.292 & 0 \\
					$8$ & 0.028 & 0.333 & 0 \\
					$9$ & 0.042 & 0.375 & 0.125 \\
					$10$ & 0.059 & 0.417 & 0.25 \\
					$11$ & 0.082 & 0.458 & 0.375 \\
					\toprule
				\end{tabular}
			}
			\caption{Weak attacks.}
		}
	\end{subtable}
\end{table*}

\subsection{Proof of Theorem \ref{theorem:aspis_optimal_attack}}
\label{appendix:fixed_diagreement_optimality}

For a given file $F$, let $A' \subseteq A$ with $|A'| \geq r'$ be the set of ``active adversaries'' in it, i.e., $A' \subseteq F$ consists of Byzantines that collude to create a majority that distorts the gradient on it. In this case, the remaining workers in $F$ belong to $\cap_{i \in A'} D_i$, where we note that $|\cap_{i \in A'} D_i| \leq q$. Let $X_j, j = r',r'+1, \dots,r$ denote the subset of files where the set of active adversaries is of size $j$; note that $X_j$ depends on the disagreement sets $D_i, i = 1,2, \dots, q$. Formally,
\begin{eqnarray}
	X_j &=& \{F: \exists A' \subseteq A \cap F, |A'| = j,\nonumber\\
	&&\qquad\text{~and~} \forall~ U_j \in F \setminus A', U_j \in \cap_{i \in A'} D_i\}. \label{eq:X_j_files}\quad
\end{eqnarray}
Then, for a given choice of disagreement sets, the number of files that can be corrupted is given by $|\cup_{j=r'}^r X_j|$. We obtain an upper bound on the maximum number of corrupted files by maximizing this quantity with respect to the choice of $D_i, i = 1,2, \dots, q$, i.e.,
\begin{equation}
	c_{\mathrm{max}}^{(q)} = \max\limits_{D_i, |D_i| \leq q,  i = 1,2,\dots, q} |\cup_{j=r'}^r X_j|
\end{equation}
where the maximization is over the choice of the disagreement sets $D_1,D_2,\dots,D_q$. 
With $X_j$ given in (\ref{eq:X_j_files}), assuming $q\geq r'$, the number of distorted files is upper bounded by 
\begin{align}
	|\cup_{j=r'}^r X_j| &\leq \sum_{j=r'}^r |X_j| \text{~(by the union bound).} \label{eq:union_bd}
\end{align}
For that, recall that $r'=(r+1)/2$ and that an adversarial majority of at least $r'$ distorted computations for a file is needed to corrupt that particular file. Note that $X_j$ consists of those files where the active adversaries $A'$ are of size $j$; these can be chosen in $\binom{q}{j}$ ways. The remaining workers in the file belong to $\cap_{i \in A'} D_i$ where $|\cap_{i \in A'} D_i| \leq q$. Thus, the remaining workers can be chosen in at most $\binom{q}{r-j}$ ways. It follows that
\begin{align}
	|X_j| \leq \binom{q}{j}\binom{q}{r-j}. \label{eq:upper_bd_X_j}
\end{align}
Therefore,
\begin{eqnarray}
	c_{\mathrm{max}}^{(q)} &\leq& {q \choose r'}{q \choose r-r'} + {q \choose r'+1}{q \choose r-(r'+1)}\nonumber\\
	&&+ \cdots \nonumber\\
	&&+ {q \choose r-1}{q \choose r-(r-1)} + {q \choose r}\label{eq:c_q_max_first_inequality}\\
	&=& \sum_{i=r'}^q{{q}\choose{i}}{{q}\choose{r-i}} \label{eq:optimal_c_q_max}\\
	&=& \sum_{i=0}^q{{q}\choose{i}}{{q}\choose{r-i}} - \sum_{i=0}^{r'-1}{{q}\choose{i}}{{q}\choose{r-i}}\label{eq:optimal_c_q_max_simplification_1}\\
	&=& \frac{1}{2}{2q\choose r}\label{eq:optimal_c_q_max_simplification_2}.
\end{eqnarray}
Eq. \eqref{eq:optimal_c_q_max} follows from the convention that ${n\choose k} = 0$ when $k > n$ or $k < 0$. Eq. \eqref{eq:optimal_c_q_max_simplification_2} follows from Eq. \eqref{eq:optimal_c_q_max_simplification_1} using the following observations

\begin{itemize}
	\item $\sum_{i=0}^q{{q}\choose{i}}{{q}\choose{r-i}} = \sum_{i=0}^r{{q}\choose{i}}{{q}\choose{r-i}} = {2q\choose r}$ in which the first equality is straightforward to show by taking all possible cases: $q<r$, $q=r$ and $q>r$.
	\item By symmetry, $\sum_{i=0}^{r'-1}{{q}\choose{i}}{{q}\choose{r-i}} = \sum_{i=r'}^{q}{{q}\choose{i}}{{q}\choose{r-i}} = \frac{1}{2}{2q\choose r}$.
\end{itemize}

The upper bound in Eq. \eqref{eq:c_q_max_first_inequality} is met with equality when all adversaries choose the same disagreement set, which is a $q$-sized subset of the honest workers, i.e., $D_i = D \subset H$ for $i =1, \dots,q$. In this case, it can be seen that the sets $X_j, j=r', \dots, r$ are disjoint so that \eqref{eq:union_bd} is met with equality. Moreover, \eqref{eq:upper_bd_X_j} is also an equality. This finally implies that \eqref{eq:c_q_max_first_inequality} is also an equality, i.e., this choice of disagreement sets saturates the upper bound.

It can also be seen that in this case, the adversarial strategy yields a graph $\mathbf{G}$ with multiple maximum cliques. 
To see this, we note that the adversaries in $A$ agree with all the computed gradients in $H \setminus D$. Thus, they form a clique of $M_{\mathbf{G}}^{(1)}$ of size $K-q$ in $\mathbf{G}$. Furthermore, the honest workers in $H$ form another clique $M_{\mathbf{G}}^{(2)}$, which is also of size $K-q$. Thus, the detection algorithm cannot select one over the other and the adversaries will evade detection; and the fallback robust aggregation strategy will apply.

\subsection{Experiment Setup Details}
\label{appendix:implementation_details}

\subsubsection{Cluster Setup}
\label{appendix:cluster_details}
We used clusters of $K=15$, $21$, and $25$ workers arranged in various setups within Amazon EC2. Initially, we used a PS of type \texttt{i3.16xlarge} and several workers of type \texttt{c5.4xlarge} to set up a distributed cluster; for the experiments, we adapted GPUs, \texttt{g3s.xlarge} instances were used. However, purely distributed implementations require training data to be transmitted from the PS to every single machine, based on our current implementation; an alternative approach one can follow is to set up shared storage space accessible by all machines to store the training data. Also, some instances were automatically terminated by AWS per the AWS \emph{spot instance} policy limitations;\footnote{\href{https://docs.aws.amazon.com/AWSEC2/latest/UserGuide/spot-interruptions.html}{https://docs.aws.amazon.com/AWSEC2/latest/UserGuide/spot-interruptions.html}} this incurred some delays in resuming the experiments that were stopped. In order to facilitate our evaluation and avoid these issues we decided to simulate the PS and the workers for the rest of the experiments on a single instance either of type \texttt{x1.16xlarge} or \texttt{i3.16xlarge}. We emphasize that the choice of the EC2 setup does not affect any of the numerical results in this paper since in all cases, we used a single virtual machine image with the same dependencies. Handling of the GPU floating-point precision errors have been discussed in Supplement Section \ref{appendix:gradient_equality}.

\subsubsection{Data Set Preprocessing and Hyperparameter Tuning}
The CIFAR-10 images have been normalized using standard mean and standard deviation values for the data set. The value used for momentum (for gradient descent) was set to $0.9$, and we trained for $16$ epochs in all experiments. The number of epochs is precisely the invariant we maintain across all experiments, i.e., all schemes process the training data the same number of times. The batch size and the learning rate are chosen independently for each method; the number of iterations is adjusted accordingly to account for the number of epochs. For Section \ref{sec:experiments}, we followed the advice of the authors of DETOX and chose $(K, b)=(15, 480)$ and $(K, b)=(21, 672)$ for the DETOX and baseline schemes. For Aspis, we used $(K, b)=(15, 14560)$ (32 samples per file) and $(K, b)=(21, 3990)$ (3 samples per file) for the ALIE experiments and $b=1365$ (3 samples per file) for the remaining experiments except for the FoE optimal attack $q=4$ (\emph{cf.} Figure \ref{fig:top1_fig_101}) for which $b=14560$ performed better. In Section \ref{sec:experiments_aspis+}, we used $(K,b) = (15,480)$ and $(K,b) = (25,800)$ for DETOX as well as for baseline schemes while for Aspis+ we used $(K,b) = (15,770)$ for the ALIE experiments and $(K,b) = (25,1800)$ for the constant attack experiments. In Supplement Table \ref{table:tuning}, a learning rate schedule is denoted by $(x,y)$; this notation signifies the fact that we start with a rate equal to $x$, and every $z$ iterations, we set the rate equal to $x\times y^{t/z}$, where $t$ is the index of the current iteration and $z$ is set to be the number of iterations occurring between two consecutive checkpoints in which we store the model (points in the accuracy figures). We also index the schemes in order of appearance in the corresponding figure's legend. Experiments that appear in multiple figures are not repeated in Supplement Table \ref{table:tuning} (we ran those training processes once). In order to pick the optimal hyperparameters for each scheme, we performed an extensive grid search involving different combinations of $(x,y)$. In particular, the values of $x$ we tested are 0.3, 0.1, 0.03, 0.01, 0.003, 0.001, and 0.0003, and for $y$ we tried 1, 0.975, 0.95, 0.7 and 0.5. For each method, we ran 3 epochs for each such combination and chose the one which was giving the lowest value of average cross-entropy loss (principal criterion) and the highest value of top-1 accuracy (secondary criterion).

\begin{figure*}[!ht]
	\centering
	\begin{subfigure}[b]{0.43\textwidth}
		\centering
		\includegraphics[scale=0.45]{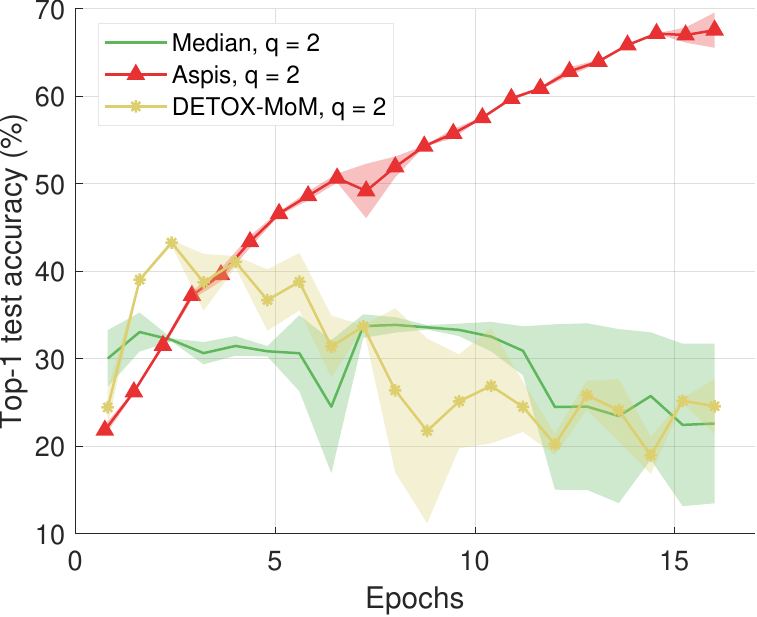}
		\caption{$q=2$ adversaries.}
		\label{fig:top1_error_bar_1}
	\end{subfigure}
	\hspace{0.07\textwidth}
	\begin{subfigure}[b]{0.43\textwidth}
		\centering
		\includegraphics[scale=0.45]{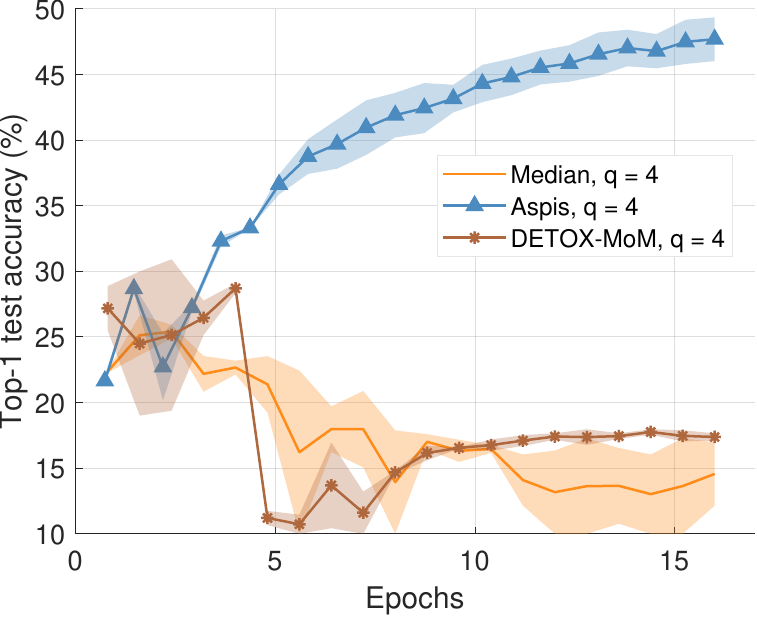}
		\caption{$q=4$ adversaries.}
		\label{fig:top1_error_bar_4}
	\end{subfigure}
	\caption{\emph{ALIE} optimal attack and median-based defenses (CIFAR-10), $K=15$ with different random seeds, ATT-2 (Aspis).}
	\label{fig:top1_error_bar_median}
\end{figure*}

\begin{figure*}[!ht]
	\centering
	\begin{subfigure}[b]{0.43\textwidth}
		\centering
		\includegraphics[scale=0.45]{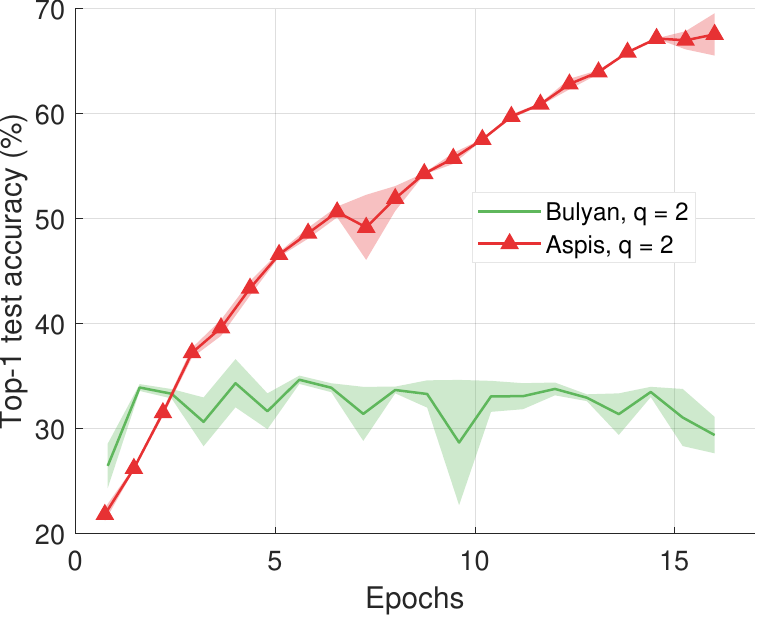}
		\caption{$q=2$ adversaries.}
		\label{fig:top1_error_bar_2}
	\end{subfigure}
	\hspace{0.07\textwidth}
	\begin{subfigure}[b]{0.43\textwidth}
		\centering
		\includegraphics[scale=0.45]{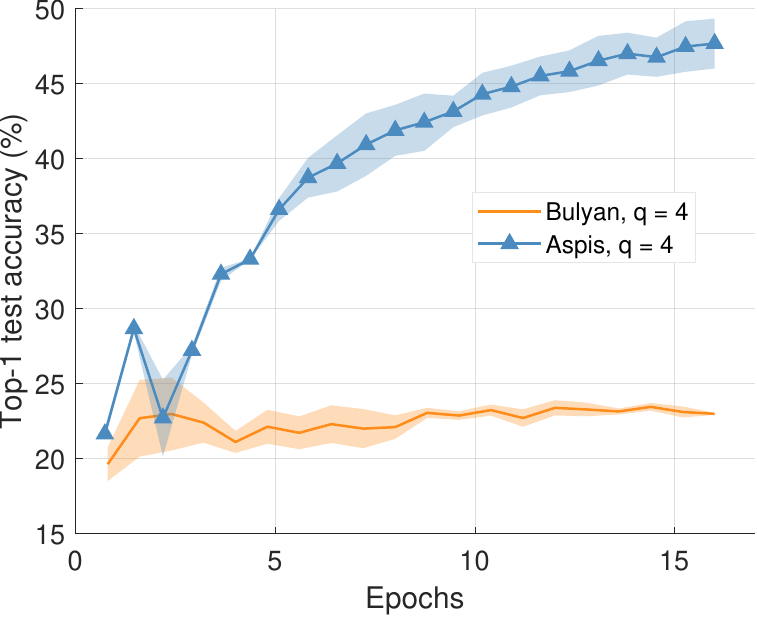}
		\caption{$q=4$ adversaries.}
		\label{fig:top1_error_bar_5}
	\end{subfigure}
	\caption{\emph{ALIE} optimal attack and \emph{Bulyan}-based defenses (CIFAR-10), $K=15$ with different random seeds, ATT-1 (Aspis).}
	\label{fig:top1_error_bar_bulyan}
\end{figure*}

\subsubsection{Error Bars}
\label{appendix:error_bars}
\begin{figure*}[!ht]
	\centering
	\begin{subfigure}[b]{0.43\textwidth}
		\centering
		\includegraphics[scale=0.45]{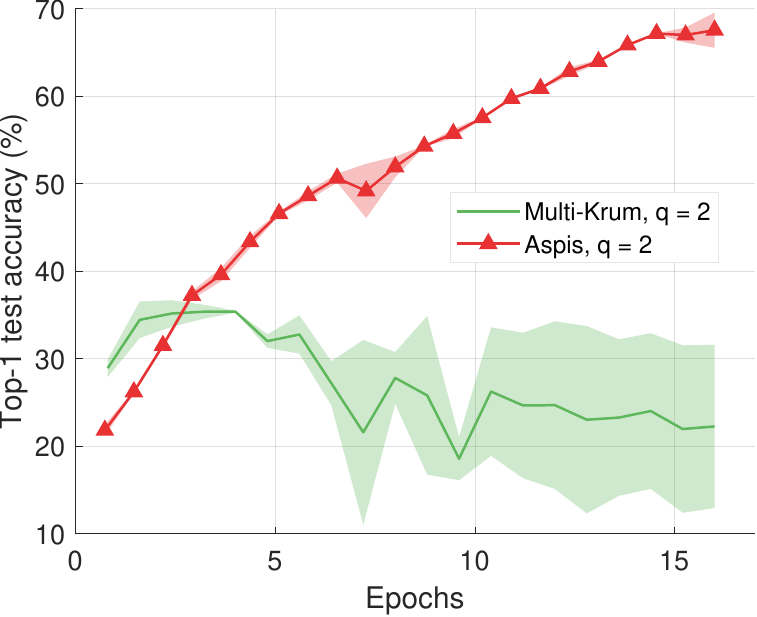}
		\caption{$q=2$ adversaries.}
		\label{fig:top1_error_bar_3}
	\end{subfigure}
	\hspace{0.07\textwidth}
	\begin{subfigure}[b]{0.43\textwidth}
		\centering
		\includegraphics[scale=0.45]{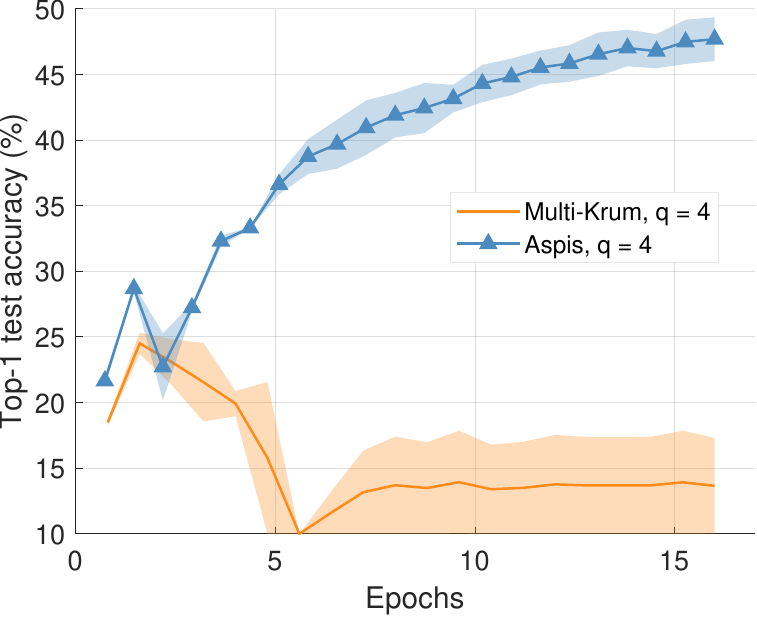}
		\caption{$q=4$ adversaries.}
		\label{fig:top1_error_bar_6}
	\end{subfigure}
	\caption{\emph{ALIE} optimal attack and \emph{Multi-Krum}-based defenses (CIFAR-10), $K=15$ with different random seeds, ATT-2 (Aspis).}
	\label{fig:top1_error_bar_multikrum}
\end{figure*}

In order to examine whether the choice of the random seed affects the accuracy of the trained model we have performed the experiments of Section \ref{sec:experiments} for the ALIE distortion for two different seeds for the values $q=2,4$ for every scheme; we used $428$ and $50$ as random seeds. These tests have been performed for the case of $K=15$ workers. In Figure \ref{fig:top1_error_bar_1}, for a given method, we report the minimum accuracy, the maximum accuracy, and their average for each evaluation point. We repeat the same process in Figures \ref{fig:top1_error_bar_2} and Supplement Figure \ref{fig:top1_error_bar_3} when comparing with Bulyan and Multi-Krum, respectively. The corresponding experiments for $q=4$ are shown in Figures \ref{fig:top1_error_bar_4}, \ref{fig:top1_error_bar_5}, and Supplement Figure \ref{fig:top1_error_bar_6}.

Given the fact that these experiments take a significant amount of time and that they are computationally expensive, we chose to perform this consistency check for a subset of our experiments. Nevertheless, these results indicate that prior schemes \cite{detox, aggregathor, bulyan} are sensitive to the choice of the random seed and demonstrate an unstable behavior in terms of convergence. In all of these cases, the achieved value of accuracy at the end of the 16 epochs of training is small compared to Aspis. On the other hand, the accuracy results for Aspis are almost identical for both choices of the random seed.

\subsubsection{Computation and Communication Overhead}
\label{appendix:computation_communication_overhead}
Our schemes provide robustness under powerful attacks and sophisticated distortion methods at the expense of increased computation and communication time. Note that each worker has to perform $l$ forward/backward propagation computations and transmit $l$ gradients per iteration. In related baseline \cite{bulyan,blanchard_krum} and redundancy-based methods \cite{detox, draco}, each worker is responsible for a single such computation. Experimentally, we have observed that Aspis needs up to $5\times$ overall training time compared to other schemes to complete the same number of training epochs. We emphasize that the training time incurred by each scheme depends on a wide range of parameters, including the utilized defense, the batch size, and the number of iterations, and can vary significantly. Our implementation supports GPUs, and we used NVIDIA CUDA \cite{cuda} for some experiments to alleviate a significant part of the overhead; however, a detailed time cost analysis is not an objective of our current work. Communication-related algorithmic improvements are also worth exploring. Finally, our implementation natively supports resuming from a checkpoint (trained model) and hence, when new data becomes available, we can only use that data to perform more training epochs. 
\subsubsection{Software}
\label{appendix:software}
Our implementation of the Aspis and Aspis+ algorithms used for the experiments builds on ByzShield's \mbox{\cite{byzshield}} PyTorch skeleton 
and has been provided along with dependency information and instructions \footnote{\href{https://github.com/kkonstantinidis/Aspis}{https://github.com/kkonstantinidis/Aspis}}. The implementation of ByzShield is available at \cite{byzshield_github} and uses the standard Github license. We utilized the NetworkX package \cite{networkx} for the clique-finding; its license is 3-clause BSD. The CIFAR-10 data set \cite{cifar10} comes with the MIT license; we have cited its technical report, as required.
\ifCLASSOPTIONcaptionsoff
  \newpage
\fi




\bibliographystyle{IEEEtran}
\bibliography{IEEEabrv,./citations}

\nocite{boyer_majority}

\cleardoublepage
\setcounter{page}{1}
\section{Supplement}
\subsection{Asymptotic Complexity}
\label{appendix:asymptotic}
If the gradient computation has linear complexity (assuming $\mathcal{O}(1)$ cost for the gradient computation with respect to one model parameter) and since each worker is assigned to $l$ files of $b/f$ samples each, the gradient computation cost at the worker level is $\mathcal{O}((lb/f)d)$ ($K$ such computations in parallel). In our schemes, however, $b$ is a constant multiple of $f$, and in general $r < l$ ($l = \binom{K-1}{r-1}$ for Aspis while $r=3$ is a typical redundancy value used in literature as well as in Aspis+); hence, the complexity becomes $\mathcal{O}(ld)$ which is similar to other redundancy-based schemes in \cite{byzshield,detox,draco}. For Aspis, as there are $\binom{K}{2}$ each sharing $\binom{K-2}{r-2}$ files, the complexity to determine their agreements and form the graph is $\mathcal{O}( \binom{K}{2}\binom{K-2}{r-2})$. The clique-finding problem that follows as part of Aspis detection is NP-complete. However, our experimental evidence suggests that for the kind of graphs we construct, this computation takes an infinitesimal fraction of the execution time. The NetworkX package \cite{networkx}, which we use for enumerating all maximal cliques, is based on the algorithm of \cite{tomita_clique} and has asymptotic complexity $\mathcal{O}(3^{K/3})$. We provide extensive simulations of the graph construction and clique enumeration time under the Aspis file assignment for $K = 50, r = 5$ and $K = 200, r = 3$ (\emph{cf.} Supplement Tables \ref{table:clique_time_weakK50}, \ref{table:clique_time_optimalK50}, \ref{table:clique_time_weakK200}, and \ref{table:clique_time_optimalK200} for the weak (ATT-1) and optimal (ATT-2) attack as introduced in Sections \ref{sec:all_attacks} and \ref{sec:detection}). We emphasize that this value of $K$ exceeds by far the typical values of $K$ of prior work, and the number of servers would suffice for most challenging training tasks. Even in this case, the cost of enumerating all cliques is negligible. For this experiment, we used an EC2 instance of type \texttt{i3.16xlarge}. The complexity of robust aggregation varies significantly depending on the operator. For example, majority voting can be done in time, which scales linearly with the number of votes using \emph{MJRTY} proposed in \cite{boyer_majority}. In our case, this is $\mathcal{O}(Kd)$ as the PS needs to use the $d$-dimensional input from all $K$ machines. Krum \cite{blanchard_krum}, Multi-Krum \cite{blanchard_krum} and Bulyan \cite{bulyan}, are applied to all $K$ workers by default and require $\mathcal{O}(K^2(d+\mathrm{log}K))$.

\subsection{Floating-Point Precision and Gradient Equality Check}
\label{appendix:gradient_equality}


A gradient equality check is needed to determine whether two gradient vectors, e.g., $\mathbf{a}$ and $\mathbf{b}$, are equal for our majority voting procedure to work. This check can be performed on an element-by-element basis or using the norm of the difference. There are two distinct cases we have considered:
\begin{itemize}
	\item \emph{Case 1: Execution on CPUs}: If we use the CPUs of the workers to compute the gradients, we have observed that two ``honest'' gradients, $\mathbf{a}$ and $\mathbf{b}$, will always be exactly equal to each other element-wise. In this case, we use the \texttt{numpy.array\_equal} function for all equality checks. If one of $\mathbf{a}$, $\mathbf{b}$ is corrupted and the other one is ``honest,'' the program will effectively flag this as a disagreement between the corresponding workers.
	\item \emph{Case 2: Execution on GPUs}: Most deep learning libraries \cite{tensorflow,pytorch} provide non-deterministic back-propagation for the sake of faster and more efficient computations. In our implementation, we use NVIDIA CUDA \cite{cuda}; hence, two ``honest'' float (e.g., \texttt{numpy.float\_32}) gradients $\mathbf{a}$ and $\mathbf{b}$ computed by two different GPUs will not be exactly equal to each other. However, the floating-point precision errors were less than $10^{-6}$ in all of our experiments. In this case, we decide that the two workers agree with each other if the following criterion is satisfied for a small tolerance value of $10^{-5}$
	$$\frac{\lVert \mathbf{a} - \mathbf{b} \rVert_2}{\mathrm{max}\{\lVert\mathbf{a}\rVert_2, \lVert\mathbf{b}\rVert_2\}} \leq 10^{-5}.$$
	On the other hand, if one of $\mathbf{a}$, $\mathbf{b}$ is distorted even by the most sophisticated inner manipulation attack ALIE \cite{alie}, then $\frac{\lVert \mathbf{a} - \mathbf{b} \rVert_2}{\mathrm{max}\{\lVert\mathbf{a}\rVert_2, \lVert\mathbf{b}\rVert_2\}}$ is at least five orders of magnitude larger and typically ranges in $[1,100]$.
\end{itemize}
In both cases, we have an integrity check in place to throw an exception if two ``honest'' gradients for the same task violate this criterion. We have not observed any violation of this in any of our exhaustive experiments.

\begin{algorithm}[!t]
	\KwIn{Loss vectors $\mathbf{l}_i$, $i = 1,2,\dots,c$, maximum iterations $T$.}
	{
		\abovedisplayskip=0pt
		\belowdisplayskip=0pt
        Set $\mathbf{l}_{\mathrm{MC}}$ to be an empty vector.\\
		\For{$t = 1$ to $T$}{
			Let $\mathbf{v}_t$ (of length $v$) be the vector with the losses of all $v$ simulations that ran for at least $t$ iterations, i.e.,
            $$\mathbf{v}_t = \begin{bmatrix}\mathbf{l}_{1,t}, \mathbf{l}_{2,t}, \dots, \mathbf{l}_{v,t}\end{bmatrix}$$
			Append $\frac{\sum_i \mathbf{v}_{t,i}}{v}$ to $\mathbf{l}_{\mathrm{MC}}$.
		}
        Return $\mathbf{l}_{\mathrm{MC}}$.
	}
	\caption{Average Monte Carlo loss across the simulations.}
	\label{alg:monte_carlo_loss}
\end{algorithm}

\subsection{Note on Computation of Average Monte Carlo Loss}
\label{appendix:monte_carlo_loss}
As discussed in Section \ref{sec:linear_regression_dist_training}, we ran 100 Monte Carlo simulations for each scheme and value of $q$. Among the 100 simulations of a given experiment we kept only those that converged to final empirical loss less than $0.1$ (\emph{cf.} Section \ref{sec:linear_regression_dist_training}). In order to report the average loss in Figures \ref{fig:loss_fig_127}, \ref{fig:loss_fig_128}, \ref{fig:loss_fig_129}, \ref{fig:loss_fig_130}, and \ref{fig:loss_fig_126} of the converged simulations we need to take into account the fact that each of them may have run for different number of iterations until convergence. To that end, let us collect the loss for each Monte Carlo simulation to a vector $\mathbf{l}_i$, $i = 1,2,\dots,c$ where $c$ is the number of converged simulations. The vectors $\mathbf{l}_i$, $i = 1,2,\dots,c$ do not necessarily have the same length; we used Algorithm \ref{alg:monte_carlo_loss} to compute and report the average loss of these vectors.

\begin{table}[!t]
	\centering
	\caption{Parameters used for training.}
	\label{table:tuning}
		\begin{tabular}{P{0.9cm}P{1.3cm}P{3cm}}
			\hline
			Figure & Schemes & Learning rate schedule \\
			\hline
			\ref{fig:top1_fig_94} & 1,2,5,6 & $(0.01,0.7)$\\
			\ref{fig:top1_fig_94} & 3,4 & $(0.1,0.95)$\\
			\ref{fig:top1_fig_95} & 1 & $(0.001,0.95)$\\
			\ref{fig:top1_fig_96} & 1,2 & $(0.01,0.7)$\\
			\ref{fig:top1_fig_97} & 1,2 & $(0.1,0.7)$\\
			\ref{fig:top1_fig_97} & 3,4 & $(0.1,0.95)$\\
			\ref{fig:top1_fig_97} & 5,6 & $(0.01,0.7)$\\
			\ref{fig:top1_fig_98} & 1 & $(0.1,0.7)$\\
			\ref{fig:top1_fig_99} & 1,2 & $(0.01,0.975)$\\
			\ref{fig:top1_fig_101} & 1,2 & $(0.1,0.7)$\\
			\ref{fig:top1_fig_101} & 3,4 & $(0.1,0.95)$\\
			\ref{fig:top1_fig_101} & 5,6 & $(0.01,0.95)$\\
			\ref{fig:top1_fig_100} & 1 & $(0.1,0.95)$\\
			\ref{fig:top1_fig_100} & 2 & $(0.01,0.7)$\\
			\ref{fig:top1_fig_88} & 1 & $(0.01,0.7)$\\
			\ref{fig:top1_fig_88} & 2 & $(0.1,0.95)$\\
			\ref{fig:top1_fig_88} & 3 & $(0.01,0.7)$\\
			\ref{fig:top1_fig_102} & 1,2 & $(0.01,0.7)$\\
			\ref{fig:top1_fig_102} & 3 & $(0.1,0.95)$\\
			\ref{fig:top1_fig_103} & 2 & $(0.01,0.7)$\\
			\ref{fig:top1_fig_104} & 2 & $(0.01,0.95)$\\
			\ref{fig:top1_fig_111} & 1,2 & $(0.01,0.7)$\\
			\ref{fig:top1_fig_111} & 3,4 & $(0.01,0.975)$\\
			\ref{fig:top1_fig_111} & 5,6 & $(0.1,0.975)$\\
			\ref{fig:top1_fig_112} & 1,2,3,4 & $(0.0003,0.7)$\\
			\ref{fig:top1_fig_112} & 5,6 & $(0.3,0.975)$\\
			\hline
		\end{tabular}
\end{table}


\begin{table*}[t]
	\large
	\newcommand\Ksubtablewidth{0.38\linewidth}
	\newcommand\Kresizetabular{\columnwidth}
	\centering
	\captionsetup[subtable]{position = below}
	\caption{Graph contruction (in seconds) and clique enumeration (in milliseconds) time in Aspis graph of $K=50$ vertices and redundancy $r=5$.}	
	\begin{subtable}{\Ksubtablewidth}
		\centering
		{
			\resizebox{0.95\Kresizetabular}{!}{
				\begin{tabular}{P{0.9cm}P{3cm}P{3cm}}
					\hline
					$q$ & Graph construction (s) & Clique finding (ms)\\
					\hline
					5 & 15 & 1 \\
					\hline
					10 & 15 & $<1$ \\
					\hline
					15 & 14 & 1 \\
					\hline
					20 & 14 & 1 \\
					\hline
				\end{tabular}
			}
			\caption{Adversaries carry out weak attack ATT-1.}
			\label{table:clique_time_weakK50}
		}
	\end{subtable}
	\hspace{0.07\textwidth}
	\begin{subtable}{\Ksubtablewidth}
		\centering
		{
			\resizebox{0.95\Kresizetabular}{!}{
				\begin{tabular}{P{0.9cm}P{3cm}P{3cm}}
					\hline
					$q$ & Graph construction (s) & Clique finding (ms)\\
					\hline
					5 & 16 & 2 \\
					\hline
					10 & 14 & 1 \\
					\hline
					15 & 14 & 1 \\
					\hline
					20 & 14 & 1 \\
					\hline
				\end{tabular}
			}
			\caption{Adversaries carry out optimal attack ATT-2.}
			\label{table:clique_time_optimalK50}
		}
	\end{subtable}
\end{table*}

\begin{table*}[t]
	\large
	\newcommand\Ksubtablewidth{0.38\linewidth}
	\newcommand\Kresizetabular{\columnwidth}
	\centering
	\captionsetup[subtable]{position = below}
	\caption{Graph contruction (in seconds) and clique enumeration (in milliseconds) time in Aspis graph of $K=100$ vertices and redundancy $r=3$.}	
	\begin{subtable}{\Ksubtablewidth}
		\centering
		{
			\resizebox{0.95\Kresizetabular}{!}{
				\begin{tabular}{P{0.9cm}P{3cm}P{3cm}}
					\hline
					$q$ & Graph construction (s) & Clique finding (ms)\\
					\hline
					5 & 4 & 51 \\
					\hline
					10 & 4 & 46 \\
					\hline
					15 & 4 & 43 \\
					\hline
					20 & 4 & 40 \\
					\hline
				\end{tabular}
			}
			\caption{Adversaries carry out weak attack ATT-1.}
			\label{table:clique_time_weakK200}
		}
	\end{subtable}
	\hspace{0.07\textwidth}
	\begin{subtable}{\Ksubtablewidth}
		\centering
		{
			\resizebox{0.95\Kresizetabular}{!}{
				\begin{tabular}{P{0.9cm}P{3cm}P{3cm}}
					\hline
					$q$ & Graph construction (s) & Clique finding (ms)\\
					\hline
					5 & 4 & 55 \\
					\hline
					10 & 4 & 54 \\
					\hline
					15 & 4 & 55 \\
					\hline
					20 & 4 & 55 \\
					\hline
				\end{tabular}
			}
			\caption{Adversaries carry out optimal attack ATT-2.}
			\label{table:clique_time_optimalK200}
		}
	\end{subtable}
\end{table*}

\end{document}